\documentclass[journal]{IEEEtran}

\usepackage{amsmath}
\usepackage{algorithm}
\usepackage{algorithmic}
\usepackage{array}
\usepackage{color}
\usepackage{graphicx}
\usepackage{url}
\usepackage{enumerate}
\usepackage{cite}

\newcommand{\tabincell}[2]{\begin{tabular}{@{}#1@{}}#2\end{tabular}}

\begin{document}

\title{JigsawNet: Shredded Image Reassembly using Convolutional Neural Network and Loop-based Composition}
\author{Canyu Le$^1$ and Xin Li$^{*2}$
\thanks{$^1$ Canyu Le is with the Department
		of Information and Science, Xiamen University, China.
		E-mail: lecanyu@gmail.com.}
\thanks{$^2$ Xin Li is with School of Electrical Engineering and Computer Science, Louisiana State University, USA. 
		E-mail: xinli@lsu.edu.}
\thanks{Manuscript received September 10, 2018.}}

\markboth{Submitted to arxiv.org}%
{Le and Li}
\maketitle

\begin{abstract}
This paper proposes a novel algorithm to reassemble an arbitrarily shredded image to its original status. 
Existing reassembly pipelines commonly consist of a local matching stage and a global compositions stage. 
In the local stage, a key challenge in fragment reassembly is to reliably compute and identify correct pairwise matching, for which most existing algorithms use handcrafted features, and hence, cannot reliably handle complicated puzzles. We build a deep convolutional neural network to detect the compatibility of a pairwise stitching, and use it to prune computed pairwise matches.
To improve the network efficiency and accuracy, we transfer the calculation of CNN to the stitching region and apply a boost training strategy.  
In the global composition stage, we modify the commonly adopted greedy edge selection strategies to two new loop closure based searching algorithms.  
Extensive experiments show that our algorithm significantly outperforms existing methods on solving various puzzles, especially those challenging ones with many fragment pieces. 
Data and code have been made available in \url{https://github.com/Lecanyu/JigsawNet}.   
\end{abstract}


%
\IEEEpeerreviewmaketitle

\section{Introduction}
\label{sec:introduction}

\IEEEPARstart{R}{eassembling} and restoring original information from fragmented visual data is essential in many forensic and archaeological tasks. In the past decades, research progress has been made in reassembling various types of fragments including 2D data such as images \cite{liu2011automated,zhang2014graph}, frescoes~\cite{brown2008system}, and 3D objects like ancient relics~\cite{son2013axially, dellepiane2011reassembling} and damaged skeletal remains \cite{GVC:Yin11ICCV,Zhang15ICCV}. 
These research works could potentially save human being from tedious and time-consuming manual composition in the restoration of valuable documents/objects/evidences in variety of practical cases.

Fragments reassembly problem can be formulated as solving a arbitrarily-cut jigsaw puzzle. Teaching computers to reliably do this, however, remains challenging, since it was first discussed in \cite{freeman1964apictorial} in 1964. 
The difficulty comes from both the local and global aspects of puzzle solving. 
(1) Locally, we need to identify adjacent pieces and correctly align them. But correlated fragments only share matchable geometry and texture along the fractured boundary. Unlike partial matching studied in classic problems such as image panorama and structure-from-motion, where the overlaps (repeated patterns) are often more significant, here the correlation between adjacent pieces is weak and difficult to identify. 
(2) Globally, even with a well-designed pairwise alignment algorithm, due to various noise and ambiguity (to be elaborated in Section~\ref{Sec:GlobalDetail}), it is usually not always reliable. Effective composition needs to take mutual consistency into account from a more global aspect. A powerful global composition algorithm, unfortunately, is often complex, computationally expensive, and prone to local optima.

To tackle the above two challenges is non-trivial. 
For \emph{local matching}, after a pairwise alignment is computed, reliably identifying whether such an alignment is correct is not easy.  
Intuitively, smooth transitions in image contexts across the fractured boundary can be a key criterion in formulating or evaluating pairwise alignment compatibility.
However, such a smoothness does not simply mean a color or gradient similarity, but is abstract and difficult to model in closed forms. 
Second, the non-smoothness also often exists in the content of an image near foreground/background contours, silhouettes, or between neighboring objects in the scene. 
Third, on regions without rich textures (e.g., pure-color backgrounds, or night skies), alignments could have great ambiguity, and incorrect stitching may also produce natural transitions in such cases. 

For \emph{global composition}, when the local pairwise alignments are unreliable (e.g. many incorrect alignments mix with correct ones), finding all the correct ones by maximizing groupwise mutual consistency is essentially an NP-hard problem \cite{demaine2007jigsaw}. An efficient and effective strategy is needed to handle complicated puzzles. 

In this work, our main idea and technical contribution in tackling these difficulties are as follows. 
Locally, we design a Convolutional Neural Network (CNN) to learn implicit image features from fragmented training data, to judge the likelihood of a local alignment being correct. 
Globally, we generalize and apply the loop-closure constraints, which have been effectively used on SLAM~\cite{mur2017orb}, environment reconstruction fields \cite{choi2015indoor, le2018sparse3d}, and previous square-shaped jigsaw puzzles \cite{son2014solving, son2016solving}, to the composition of arbitrarily shredded (geometrically irregular) image fragments. 

In summary, the main contributions of this work are 
\begin{itemize}
	\item We design a CNN network to evaluate the pairwise compatibility between fragment pairs. 
	To improve the network performance, two technical components are designed: (1) the transfer of the CNN calculation attention on stitching regions, and (2) an adaptive boosting training procedure for solving the data imbalance problem. 
	\item We develop a new loop-closure based composition strategy to enforce mutual consistency among poses of multiple pieces. This greatly improves the robustness of global composition, especially in solving complex puzzles. 
\end{itemize}

We have conducted thorough experiments on various benchmarks. Our approach greatly outperforms existing state-of-the-art methods in puzzle solving.
Codes and data have been released to facilitate future comparative study on image reassembly and related research.

\section{Related Works}
Originated from Freeman et al. \cite{freeman1964apictorial}, the jigsaw puzzle solving problem has been exploited in many literatures.
Generally, we can categorize this problem into solving regular shape puzzles and solving irregular shape puzzles.

\subsection{Solving Regular-Shaped Jigsaw Puzzles}
Square jigsaw puzzles are the most typical cases in regular shape jigsaw puzzle. 
Recently, multiple literatures have studied this problem. 
Cho et al. \cite{cho2010probabilistic} evaluate inter-fragment consistency using the \emph{sum-of-squared} color difference (SSD) alone the stitching boundary, and used a graphical model to solve the global composition.
Pomeranz et al. \cite{pomeranz2011fully} exploit various measurement strategies to improve the accuracy of pairwise alignment compatibility, and also introduced a consensus metric to the greedy solver in global composition. 
Gallagher et al. \cite{gallagher2012jigsaw} develop a \emph{Mahalanobis Gradient Compatibility} (MGC) to evaluate the pairwise alignment using changes in intensity gradients, rather than changes in intensity itself near the boundary; in the global composition stage, they greedily generate a minimal spanning tree to connect all the pieces.
More recently, state-of-the-art square jigsaw puzzle solving results were reported in \cite{son2014solving} and \cite{son2016solving}. 
In \cite{son2014solving}, Son et al. exploit the loop constraints configuration to filter out false negative alignments; later in \cite{son2016solving}, the aforementioned MGC measurement is improved by a more accurate intensity gradient calculation, and the overall reassembly is further enhanced by improving the consensus composition.

Those state-of-the-art square puzzle solvers can process even more than a thousand fragments. However, square solvers cannot be used to handle general puzzles that have arbitrary shaped fragments. 
The key \textbf{difference} is on the assumption of fragmented pieces being square. Such a simplification makes this problem combinatorial: fragments always locate in a 2D array of cells indexed by a pair of grid integers $(i,j)$, and the rotation is just $k \times \pi/2$. On such square fragments, pairwise compatibility measurement, such as SSD, MGC, and its variants, can simply consider pixel intensity/gradient consistency along horizontal and vertical directions on the straight boundary. 
From the global aspect, loop closures can be easily formulated and detected on a 2D grid. 
Algorithms developed based on these simplifications will not work on general puzzles.

A CNN-based method was explored in \cite{paumard2018jigsaw} recently. Paumard et al. designed a neural network to predict fragments relative position, and then a greedy strategy is applied for global composition. However, their method can only tackle the simple puzzles and the number of pieces they solved in their experiments is nine.

\subsection{Solving Irregular-Shaped Jigsaw Puzzles}
Irregular-shaped jigsaw puzzles are composed of arbitrarily cut fragmented pieces. 
Shredded images or documents are typical and practical cases of such puzzles.

Color information from the boundary pixels was used in building image fragment descriptors for their matching. 
Amigoni et al.~\cite{Amigoni03ICIP} extract color content from the fragments' boundary outlines, and use them to match and align image pieces. 
Tsamoura et al. \cite{tsamoura2010automatic} apply a color-based image retrieval strategy to identify potential adjacent fragments, and then use boundary pixel's color to build the contour feature. The pairwise matching is then computed by finding a longest common subsequence~\cite{Wolfson90PAMI} between fragments' contours. 
In \cite{Sagiroglu06ICPR}, the texture of a band outside the border of pieces is predicted by image inpainting. An FFT-based registration algorithm is then utilized to find the alignment of the fragment pieces.

Fragment's boundary geometry is also commonly used in building features for fragment matching. Zhu et al.~\cite{zhu2008globally} approximate contours of ripped pieces by polygons and use the turning angles defined on the polygons as the geometric feature to match fragments.  
Liu et al. \cite{liu2011automated} also use polygons to approximate the noisy fragment contours and then extract vertex and line features along the simplified boundary to match partial curves. Each pairwise matching candidate contains a score to indicate how well the matching is. They use those scores to build a weighted graph and apply a spectral clustering technique to filter out irrelevant matching. 
Zhang et al. \cite{zhang2014graph} build the polygon approximation on both the geometry and color space, and use ICP to compute potential pairwise matches. Multiple pairwise alignments are stored on a multi-graph, weighted by pairwise matching scores. The global composition is solved by finding a simple graph with maximized compatible edge set through a greedy search. 

All these existing puzzle solvers generally follow the a three-step composition procedure: 
(1) design geometry- or color-based features to describe the fragments; 
(2) compute the inter-fragment correspondences and/or rigid transformations (alignments) between pieces, rank these alignments using a score;
(3) globally reassemble the pieces using acceptable pairwise alignments.
However, these existing algorithms not only depend on having well designed features, but also often need parameters carefully tuned. 
This becomes very difficult in general, as puzzles could have different contents and different complexities, and a set of predetermined handcrafted features and hand-tuned parameters may not work for all the various cases.
In most experiments reported in all these existing literatures, the puzzles are relatively simple, and the fragment numbers are smaller than $30$. 

\section{Overview} 
\label{Sec:overview}

\begin{figure*}[h]
	\centering
	\includegraphics[height=0.12\textheight]{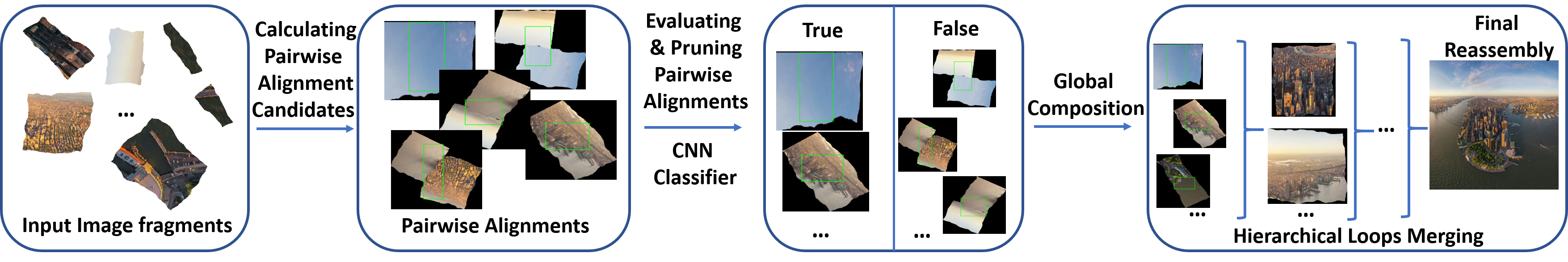}
	\caption{Image reassembly algorithm pipeline. Given the image fragments, we first calculate pairwise alignments to get many pairwise alignment candidates. Then, we use a CNN detector to classify the potentially correct alignments from the incorrect ones. Finally, we do a global composition by maximizing mutual consistency among fragments using loop closure constraints. \label{Fig:pipeline}}
\end{figure*}

As illustrated in Fig \ref{Fig:pipeline}, our approach contains three components: pairwise alignment candidates extraction, pairwise compatibility measurement, and global composition.
Intuitively, if following a pairwise matching, two fragments can align (under rigid transformation) with natural geometry and texture transition across the boundary, we consider this as a candidate alignment. 
But before matching, we do not know whether two fragments are adjacent or not. Therefore, we compute matching between every pair of fragments. 
We adopt the pairwise matching computation strategy used in \cite{zhang2014graph}, which formulates the matching as a partial curve matching problem. 
Specifically, it contains four steps: 
(1) Through a revised RDP algorithm~\cite{douglas1973algorithms}, approximate the noisy fragment boundary contour into a polygon, whose each line segment has similar color;
(2) Match each fragment pair by iteratively estimating all the possible segment-to-segment matches;
(3) Refine those good segment-to-segment matches using an ICP algorithm; and
(4) Evaluate the pairwise matching score by calculating the volume of well aligned pixels. 
Between each fragment pair, this matching algorithm produces a set of possible alignments, in which both correct and incorrect alignments exist and the number of incorrect alignments is much bigger than correct ones.

\textbf{Pairwise compatibility measurement}.
With pairwise alignment candidates, we still need a reliable compatibility evaluator that can examine these candidates: to keep ones that are probably correct and filter out ones that are likely incorrect.
Such a detector could reduce the search space in the next global composition step, and benefit both reassembly robustness and efficiency. 
In most existing reassembly algorithms, heuristic and handcrafted features and evaluation schemes are designed to measure such a compatibility. 
For example, in~\cite{zhang2014graph}, the alignment score is defined as the number of matched pixels (i.e., after the ICP transformation on fragments, pixels that have similar color, opposite normal, and small spatial distance).
In \cite{tsamoura2010automatic}, the matching score is defined as the weighted length of the extracted longest common subsequence. 
However, these manually designed evaluators do not always work well for different puzzles, and the parameter tuning is often difficult. 
Hence, in this work, we design a pairwise compatibility detector (classifier) using a CNN network. This network is trained to identify whether the stitching of an image fragment pair under a specific pairwise alignment is correct or not. 

\textbf{Global Composition}.
Even with a good pairwise compatibility detector, misalignments due to local ambiguity are sometimes inevitable. 
Such errors need to be handled from a global perspective. 
We use mutual consensus of many pieces' poses to prune the pairwise alignments and globally compose the fragments. 
A widely adopted consensus constraint is \emph{loop closures}: correct pairwise alignments support each other spatially and their relative transformations compose to identity along a closed loop if they are considered in a loop. 
Such a loop closure constraints have been widely applied on refining or pruning pairwise alignments in many vision-based SLAM and environment reconstruction like  \cite{kummerle2011g,choi2015indoor,le2018sparse3d}. However, enforcing loop closures in jigsaw puzzle solving problem is more challenging than it is in these SLAM and reconstruction tasks. 
First, in reassembly, outliers dominate inliers, and furthermore, due to the significantly smaller overlap between adjacent pieces and the existence of small loops, incorrect alignments could sometimes form closed loops. Simply applying greedy loop closures, which is a common strategy in the state-of-the-art SLAM systems, will not work reliably. 
Second, besides loop closure constraints, the global composition should also prevent any inter-fragment intersection, and this extra constraint cannot be formulated in a continuous closed form together with the loop closure constraint. Enforcing it also makes the solving significantly more expensive. 
Inspired by \cite{le2018sparse3d, son2014solving}, we develop closed loops searching and merging algorithms on a general multi-graph. These algorithms have promising application on not only fragment reassembly, but also general SLAM and environment reconstruction when datasets are sparsely sampled.

\section{Pairwise Compatibility Measurement}
Pairwise matching computation results in both correct and incorrect alignments. 
Although we could develop a composition algorithm to prune the incorrect alignments using mutual consensus in the final global step, it is computationally expensive. 
When the puzzle is complex, fully relying on global pruning is prohibitive. 
An effective candidates filtering and pre-selection tool is important to both composition efficiency and reliability. 

In most existing puzzle solving algorithms, manually designed pairwise matching scores and heuristic thresholds based on experiments or parameter tuning are often used for this filtering. 
Unfortunately, building handcrafted features and weighting parameters to evaluate the stitching of various puzzles (that have different contents and geometry/size complexity) is in general very difficult. Because evaluating whether the stitched content exhibits a natural transition involves analysis in not only geometry, color, texture, but also higher-level semantics. 

Therefore, instead of handcrafted detectors, we formulate the problem of whether an alignment is correct or not as a binary classification problem, and train a CNN to do this pairwise compatibility measurement.

\subsection{A CNN Detector for Compatibility Measurement}
\label{Sec:DetectorDesign}

\subsubsection{Overview and Main Idea}
Popular convolutional neural networks architectures such as AlexNet \cite{krizhevsky2012imagenet}, VGG16 \cite{simonyan2014very} and ResNet \cite{he2016deep} have been developed and applied in many image classification and recognition tasks.  
But these classic tasks are different from fragment stitching compatibility measurement on two aspects:
(1) Instead of dealing with rectangular images that often contain relatively complete contents, in this problem, the shape of image fragment is irregular, and its content is often very local and incomplete. 
(2) In conventional classification or recognition, features from local to global, and from all over the images may contribute to classification. However, in fragments composition problem, features extracted near the stitching region which could characterize the image content transition smoothness are most important.

Therefore, we design a new CNN network, integrating the desirable properties from the structures of \emph{residual block}~\cite{he2016deep} and \emph{RoIAlign}~\cite{he2017mask}.
The intuition comes from two observations. 

First, for pairwise compatibility measurement, calculating complicated and deep feature maps is often unnecessary, because the content in a image fragment is local and incomplete. 
So unlike many other high-level recognition tasks, it does not need be built upon deep and complicated feature map stacks. 
Therefore, we build a relatively deep network ($29$ CONV layers in total, see below for detail), but make the stack of feature map shallow (with the maximum feature map stack being $128$). 
The number of parameters in such net structure are much fewer than the popular deep backbone nets such as ResNet \cite{he2016deep}, and the training (optimization iteration) and testing (evaluating) is significantly faster.

\begin{figure}[h]
	\centering
	\begin{tabular}{cc}
		\includegraphics[height=0.13\textheight]{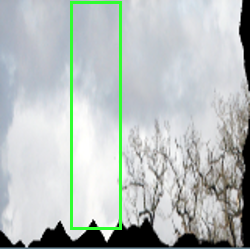} &
		\includegraphics[height=0.13\textheight]{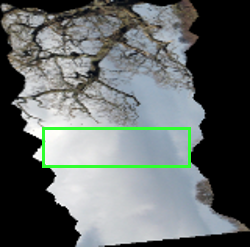} \\
	\end{tabular}
	\caption{Smooth content transition in stitching regions (green boxes). Non-smooth content transition in other regions, such as tree branches versus the background sky, is less important in evaluating the stitching compatibility. \label{Fig:RoIMotivation}}
\end{figure}	

Second, the key clues in differentiating correct and incorrect alignments should locate near the stitching boundary. 
Correct pairwise reassembly will preserve smooth transition in contents. 
A simple low-level content smoothness could be the smoothness of color intensity or smoothness of gradients (especially in regions that do not have complicated textures). 
Fig. \ref{Fig:RoIMotivation} illustrates two examples. 
In the stitching region (the green boxes), natural transitions are important.
But in other regions, less smooth transitions in contents may be ignorable. Therefore, on the one hand, the CNN should be trained to observe the smoothness of contents transition, on the other hand, the network should focus its attention on the critical stitching regions. This will not only speed the learning procedure, but also improve the recognition accuracy. 

\subsubsection{Network Architecture Design}	

\textbf{Focusing on Region of Interest (RoI).}
The region of interest alignment method (\emph{RoIAlign})~\cite{he2017mask} is applied to transfer the attention of calculation to the stitching regions. 
\emph{RoIAlign} is a pooling layer in the neural network to extract feature maps from each specific region of interest (RoI). To smoothly calculate the specific output size of this layer, a bilinear interpolation is used. 
We apply the \emph{RoIAlign} in the last pooling layer and still use the conventional \emph{max pooling} in shallow pooling layers. 
There are two benefits on this design: 
(1) The \emph{max pooling} in shallow layers can effectively increase receptive field. 
(2) In the last \emph{RoIAlign}, only features located in RoIs are calculated and the final classification (in fully connected layers) is performed mainly based on the stitching regions. 
Based on this design, the network training is performed on not only local stitching regions but also the whole image context. 
The RoIs and RoIAlign have been demonstrated in Fig. \ref{Fig:traning_net} red boxes. 

\begin{figure}[h]
	\centering
	\includegraphics[height=0.1\textheight]{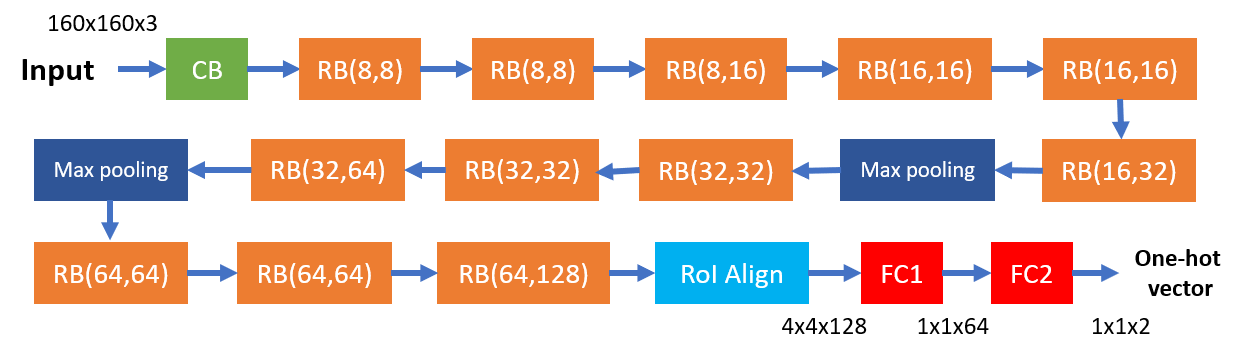}
	\caption{The convolutional neural network architecture. $CB$ is the convolutional block. $RB(r,h)$ is the residual block with depth of input $r$ and depth of output $h$. \label{Fig:net_arch}}
\end{figure}

\textbf{Network Architecture.}
The input of the neural network is a series of $160\times160\times3$ images with corresponding weights and bounding box coordinate which covers the abutted area between two images.
The original input image is processed by a convolutional block and 12 residual blocks. 
The convolutional block ($CB$) applies the following modules:
\begin{enumerate}[(1)]
	\item Convolution of 8 filters, kernel size $3\times3$ with stride 1.
	\item Batch normalization \cite{ioffe2015batch}.
	\item A rectified linear unit (ReLU).
\end{enumerate}
The residual block ($RB(r,h)$) has two parameters: the depth of input $r$ and the depth of output $h$. Each residual block has below architecture: 
\begin{enumerate}[(1)]
	\item Convolution of $h$ filters, kernel size $3\times3$ with stride 1.
	\item Batch normalization.
	\item A rectified linear unit (ReLU).
	\item Convolution of $h$ filters, kernel size $3\times3$ with stride 1.
	\item A skip connection. \\ \indent If $r=h$, then directly connect input to the block. \\ If $r\ne h$, then apply Convolution of $h$ filters of kernel size $3\times3$ with stride 1, and following batch normalization.
	\item  A rectified linear unit (ReLU).
\end{enumerate}
The output of the residual towel is passed into either a \emph{max pooling} or a \emph{RoIAlign}.
The \emph{RoIAlign} crops and resizes the feature map, which locate in the input bounding box, to $4\times4$ small feature map by using bilinear interpolation.
Finally two fully connection layers convert the feature map to the one-hot vector (i.e. a $2\times1$ vector).
Fig. \ref{Fig:net_arch} illustrates the complete network architecture.
Several experiments in Section~\ref{Sec:experiments} demonstrate the effectiveness of this new network.

\subsubsection{Training the Detector}
\textbf{Synthesizing shredded images.} 
To train our CNN detector, we build a shredding program to simulate the fragmentation of a given image.
The shredding is controlled by three parameters: (1) puzzle complexity (number of cuts to generate), (2) randomized cutting orientation, (3) perturbations along the cutting curve.  		
With this generator, we can synthesize big amount of fragmented image data for training and testing. 

\begin{figure}[h]
	\centering
	\begin{tabular}{cccc}
		\includegraphics[height=0.072\textheight]{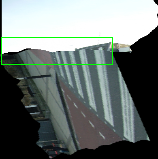} &
		\includegraphics[height=0.072\textheight]{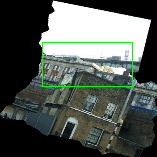} &
		\includegraphics[height=0.072\textheight]{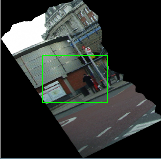} &
		\includegraphics[height=0.072\textheight]{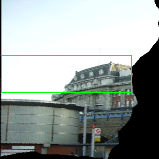} \\
		\tabincell{c}{(a) $\gamma=0.00$ \\ G.T. = False}  & 
		\tabincell{c}{(b) $\gamma=0.52$ \\ G.T. = False} & 
		\tabincell{c}{(c) $\gamma=0.78$ \\ G.T. = True} & 
		\tabincell{c}{(d) $\gamma=0.94$ \\ G.T. = True}  \\
	\end{tabular}
	\caption{Some CNN classification results. $\gamma$ is the output probability/score. G.T. stands for the groundtruth. Typically, we use a score threshold $0.5$ to distinguish correct and incorrect alignment. Here, (b) is misjudged by the CNN.  
		\label{Fig:CNNJudgement}}
\end{figure}

To train the CNN detector, first, synthesized image fragments are aligned using the aforementioned pairwise matching algorithm; these alignments are used to stitch two image fragments; then, the stitched images are fed into the CNN to train or test. 
The output of network is normalized by \emph{softmax} function which represents the \emph{probability} of true-and-false classification. We call this probability the \emph{alignment score} $\gamma$. Fig. \ref{Fig:CNNJudgement} illustrates some examples of classification results.

\subsection{Solving Data Imbalance}		
\label{Sec:DataImbalance}

\begin{figure}[h]
	\centering
	\begin{tabular}{ccc}
		\includegraphics[height=0.1\textheight]{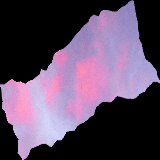} &
		\includegraphics[height=0.1\textheight]{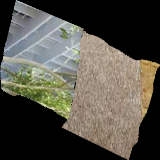}&
		\includegraphics[height=0.1\textheight]{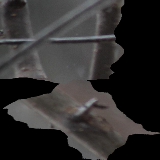} \\
		(a)  & (b)  & (c)  \\	
	\end{tabular}
	\caption{Examples of consistent and inconsistent transition. 				
		A correct pairwise alignment (a) often has consistent transition across the stitching region in both texture content and geometry. 
		The incorrect alignments usually exhibit certain inconsistency, either in texture/color content (b), or in geometry (c), along the stitching boundary.~\label{Fig:error_type}}
\end{figure}

When training the CNN detector, we compute many pairwise alignments from synthesized image fragments. 
However, these computed alignments are imbalanced. 
Between each pair of fragments, there is only one correct alignment, but the partial matching could potentially find many non-intersected alignments. 
These incorrect alignments, which indicate various non-compatible transitions, are valuable for enhancing the detector's abilities of generalization and recognition. 
We call such true-alignment versus false-alignment imbalance as \textbf{between-class imbalance}.

Furthermore, among the false alignments, there are errors due to inconsistent boundary geometry and inconsistent image contexts. 
In the alignment candidates calculation, the original fragment boundary is approximated by a polygon and the alignments (transformations) are computed by matching polygon edge pairs. Much more inconsistent geometry alignments, than inconsistent image content alignments, will be generated. 
Fig.~\ref{Fig:error_type} illustrates examples of these two types of inconsistency. 
(b) shows a typical context inconsistency in which two color-unrelated fragments were stitched although geometrically the stitching fits very well. 
(c) shows a geometry inconsistency case where the color context seems to transits well, but the stitching is not very geometrically desirable.  

Since the majority of synthesized undesirable alignments are due to geometry inconsistency, they could dominate the learning procedure. 
Also, the feature that describe image context inconsistency is harder to learn, because the image context could significantly vary from one image to another one.
Therefore, if we train a CNN directly, the detector tends to be dominated by the majority class of geometry inconsistency, and misses the detection on context inconsistency. 
The detector will achieve high overall accuracy but low precision. 
We call this type of imbalance as \textbf{within-class imbalance}.

Although data imbalance problems are related to our synthesis strategy, it is a general and fundamental issue, and is difficult to avoid and overcome by only improving synthesis strategy. 
In fact, although such geometry inconsistency versus image context inconsistency is the simple imbalance we observed, we don't know whether within each class of inconsistency, whether there are other minor sub-classes that could be dominated by other false alignments. We need a general strategy to tackle this within-class imbalance.

\subsubsection{Solving between-class imbalance}
\label{Sec:BetweenClass}
Strategies to deal with imbalanced training data for CNN can be categorized into data-level and classifier-level approaches~\cite{he2009learning, buda2017systematic}.  
\textbf{Data-level} approaches modify the original training data by either (1) \emph{oversampling}, which randomly replicates some minority classes, or (2) \emph{undersampling}, which randomly remove some majority classes.
\textbf{Classifier-level} approaches adjust the objective function or classifier accordingly. 
For example, probability distribution on imbalanced classes can be computed, then compensated by assigning different weights to different classes~\cite{elkan2001foundations,richard1991neural}.

\emph{Undersampling} data from the majority classes is an easiest approach and it could also desirably improve the efficiency of the training procedure as a smaller training dataset is considered. 
However, as various incorrect alignments are valuable in training robust detector, we find that through providing more comprehensive alignment data, \emph{oversampling} the minority class can improve the final classification accuracy. 
Furthermore, according to Buda et al. \cite{buda2017systematic}, \emph{oversampling} is generally more effective for CNN networks, and will not cause overfitting problem (which was an issue in classic machine learning models). 
Therefore, considering both the training accuracy and efficiency, we apply both \emph{oversampling} and \emph{undersampling} on the synthesized alignment datasets.  
In our experiments, we oversample the original positive datasets by $20$ times, then randomly downsample the result to a half. 
Our final training dataset contains approximate $600k$ alignments (stitched images), in which $70\%$ and $30\%$ are false and true alignments respectively.

\subsubsection{Solving within-class imbalance}
\label{Sec:WithinClass}

There is also a within-class imbalance in the data. This imbalance problem is a more difficult issue to tackle. 
As discussed in Section~\ref{Sec:DataImbalance}, during pairwise matching, much more false alignments with geometric inconsistency are generated, than false alignments with image context inconsistency. 
But we do not know which type of inconsistency exists on a specific false alignment, even with the help of groundtruth. 
Therefore, this imbalance cannot be eliminated through data over/under-sampling or re-weighting the objective function. 

The boosting methods, such as \emph{Adaboost}~\cite{freund1999short}, provides an effective mechanism in solving such imbalance. 
The boosting method combines multiple weak learners.
Each weak learner is trained on the data where the previous weak learners perform badly to complement and fortify overall result. 
Such a classifier ensemble strategy is suitable for our within-class imbalance problem. 
Mis-predicted data from the previous learners usually belong to the minority category, and these data will be assigned with a bigger weight in the next learner training. 

\textbf{Binary classification boosting}. 
Given training data $\{(x_1,y_1), (x_2,y_2), ..., (x_n,y_n)\}, x_i\in \mathcal{X}, y_i\in\mathcal{Y}=\{-1, 1\}$, to find a mapping $f: \mathcal{X}\rightarrow \mathcal{Y}$, the Boosting classifier $f$ uses several standalone \emph{learners} (i.e., weaker classifiers).
\begin{equation}
f(x) = \sum_{k=1}^{K} \alpha_k G_k(x).
\label{Eqn:assembly}
\end{equation} 
where $G_k(x)$ is the $k$-th learner, and $\alpha_k$ is the weight which measures how important this learner is in the final classifier. 
The classification error can be measured by an exponential loss function \cite{freund1999short}:
\begin{equation}
L(y, f(x)) = \exp(-yf(x)).
\label{Eqn:boost_loss}
\end{equation}
If we have a boosting classifier $f_{k-1}(x)$ from the first $k-1$ learners, finding the best $k$-th learner $G_{k}^{*}(x)$ to have a minimized loss of Eq.~(\ref{Eqn:boost_loss}) reduces to 
\begin{align}
\label{Eqn:solving}
(\alpha_k^*, G_k^*(x)) &= \arg\min_{\alpha, G} \sum_{i=1}^{n}\exp(-y_if_{k}(x_i)) \nonumber \\
&= \arg\min_{\alpha, G} \sum_{i=1}^{n}\exp\left[-y_i(f_{k-1}(x_i)+\alpha G(x_i)\right] \nonumber \\
&= \arg\min_{\alpha, G} \sum_{i=1}^{n}w_{k-1,i}\exp\left[-y_i\alpha G(x_i)\right] 
\end{align}
where $w_{k-1,i} = \exp(-y_i f_{k-1}(x_i))$. If $w_{k-1, i}$ is larger, it means the previous $k-1$ ensemble result is undesirable in data $(x_i, y_i)$. This data is assigned a heavier weight for the current learner $G_k(x)$ training. 
Therefore, $w_{k-1,i}$ can be seen as a weight distribution on the training data. The misjudged data will be amplified on the training of next learner. 

To this end, we can generally formulate $w_{k, i}$ as
\begin{equation}
\label{Eqn:w}
w_{k, i} = \exp(-y_i f_{k}(x_i)) = w_{k-1, i}\exp(-y_i\alpha_kG_k(x_i)).
\end{equation}
Eq.~(\ref{Eqn:w}) means $w_{k,i}$ is only related with the current learners ensemble. $w_{k-1,i}$ is a constant in the $k$-th calculation. 
This Eq.~(\ref{Eqn:w}) explains how to update data weight distribution $\{w_{k,1}, w_{k,2}, ..., w_{k,n}\} \in D_k$ in the $k$-th iteration according to the misclassified data.

Since $w_{k-1,i}$ is a constant for solving $G_k(x)$ and $y_i, G(x_i) \in \{-1, 1\}$, for $\forall \alpha>0$ we can separately formulate the best $G_k^*(x_i)$ in Eq.~(\ref{Eqn:solving}) as
\begin{align}
\label{Eqn:best_G}
G_k^*(x) &= \arg\min_G\sum_{i=1}^{n} w_{k-1,i}I(y_i\ne G(x_i)) \\
I(y_i\ne &G(x_i)) =
\begin{cases}
1 & \mbox{if $ y_i\ne G(x_i) $}\\
0 & \mbox{if $ y_i = G(x_i) $}
\end{cases}. \nonumber
\end{align} 
Bringing $G_k^*$ of Eq.~(\ref{Eqn:best_G}) into Eq.~(\ref{Eqn:solving}) we have

\begin{equation}
\begin{array}{l}
\label{Eqn:deriviation}
\sum_{i=1}^{n}w_{k-1,i} \exp\left[-y_i\alpha G_k(x_i)\right] \\
=  \sum_{y_i=G_k(x_i)} w_{k-1,i}e^{-\alpha} + \sum_{y_i\ne G_k(x_i)} w_{k-1,i}e^{\alpha}  \\
=  (e^{\alpha}-e^{-\alpha})\sum_{i=1}^{n} w_{k-1,i}I(y_i\ne G_k(x_i)) + e^{-\alpha}\sum_{i=1}^{n} w_{k-1,i}.
\end{array}
\end{equation}
Finally, combining Eq.~(\ref{Eqn:solving}) and Eq.~(\ref{Eqn:deriviation}), we have
\begin{equation}
\begin{array}{cl}
\label{Eqn:deriviation_best_a}
\alpha_k^* & = \arg\min_{\alpha} \sum_{i=1}^{n}w_{k-1,i}\exp\left[-y_i\alpha G_k(x_i)\right] \\
&= \arg\min_{\alpha} \left[ (e^{\alpha}-e^{-\alpha})E_k + e^{-\alpha}  \right],\\
\end{array}
\end{equation}
where $E_k = \frac{\sum_{i=1}^{n} w_{k-1,i}I(y_i\ne G_k(x_i))}{\sum_{i=1}^{n} w_{k-1,i}}$. $w_{k-1, i}$ is the classification error weight from the previous classifier $f_{k-1}(x_i)$. 
$E_k$ measures the performance of current learner $G_k(x)$ on the previous classification error weight distribution.

Finally, deriving Eq. \ref{Eqn:deriviation_best_a} with respect to $\alpha$, we get 
\begin{equation}
\label{Eqn:best_a}
\alpha_k^* = \frac{1}{2} \log\frac{1-E_k}{E_k}.
\end{equation}

Equations~(\ref{Eqn:best_G}) and (\ref{Eqn:best_a}) tell us how to optimize the current learner in the $k$-th iteration. 
Based on the above derivations, we can design the boosting algorithm for CNN training.

\textbf{CNN boost training}.
The Eq.~(\ref{Eqn:best_G}) in common boosting algorithms is discrete class tag, but the output of our CNN is continuous value. 
Therefore, to optimize the CNN, we re-formulate Eq.~(\ref{Eqn:best_G}) using \emph{cross-entropy}:
\begin{equation}
\label{Eqn:best_G_CNN}
G_k^*(x) = \arg\min_G\sum_{i=1}^{n} w_{k-1,i}\left[y_i\log\hat{y_i}+(1-y_i)\log(1-\hat{y_i}) \right] 
\end{equation}
where $y_i$ is the groundtruth and $\hat{y_i}$ is the estimation of CNN after \emph{softmax} normalization. 

\begin{figure}[h]
	\centering
	\includegraphics[height=0.2\textheight]{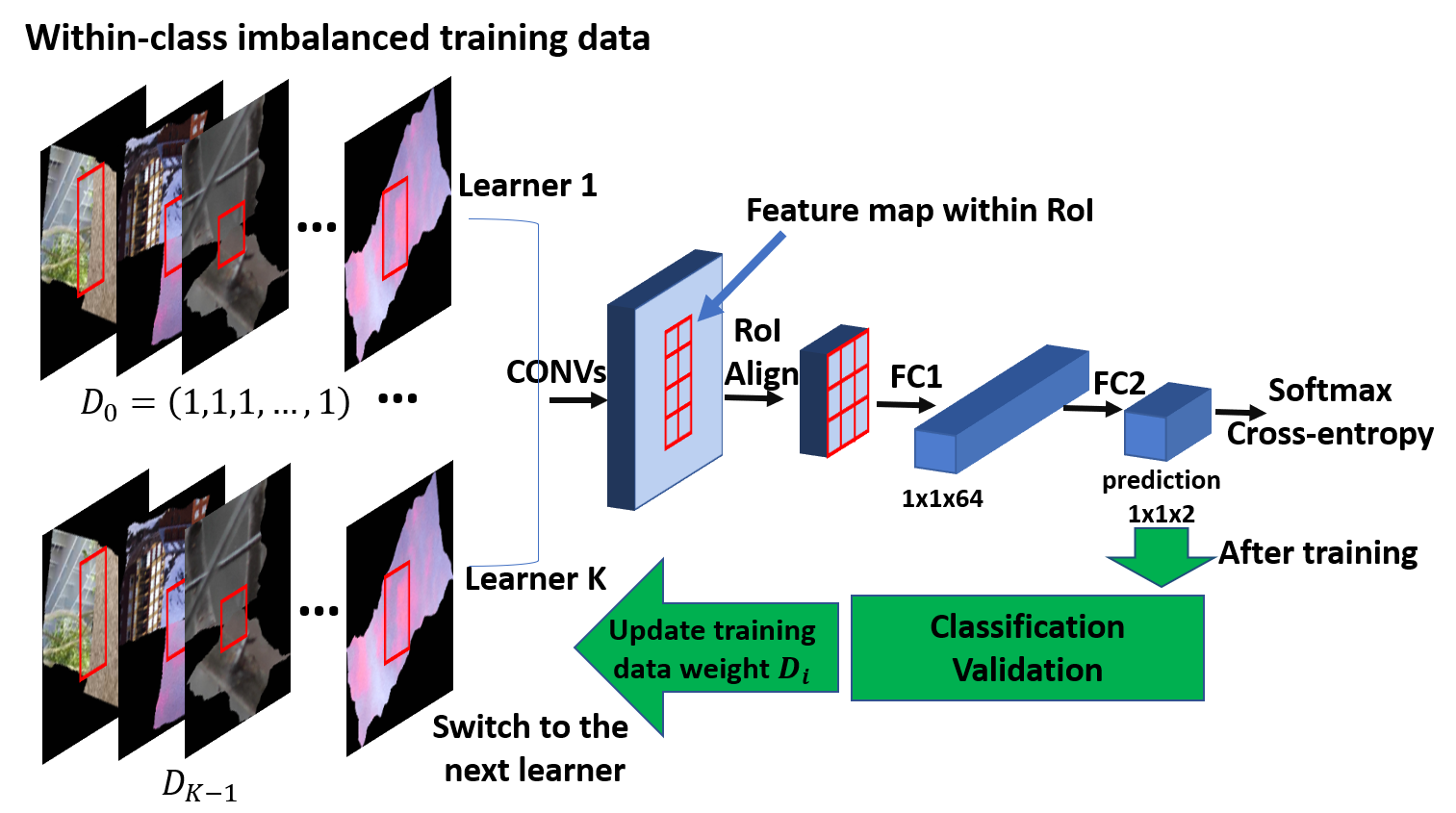}
	\caption{The CNN boost training. 
		All learners share the same network architecture, and are trained independently. 
		Each learner is trained on weighted training data from scratch.~\label{Fig:traning_net}}
\end{figure}

During training phase, the class tags in Eq. \ref{Eqn:best_G_CNN} is $\{0, 1\}$ instead of $\{-1, 1\}$. 
During validation phase, since the output of CNN is a probability of correct alignment, we can use Eq. (\ref{Eqn:convert}) to convert the probability to a discrete classification result and then apply boosting iteration without any violations.
\begin{align}
\label{Eqn:convert}
G_k^*(x) =
\begin{cases}
-1 & \mbox{if $ \hat{y_i} < p $}\\
1 & \mbox{if $ \hat{y_i} \ge p $}
\end{cases}
\end{align}  
where $p$ is a probability threshold (we set $p=0.5$ in all of our experiments).  
The entire training procedure can be summarized in Algorithm~\ref{Alg:LocalReassembly}, 
and the whole training procedure is illustrated in Fig. \ref{Fig:traning_net}.

\begin{algorithm}  
	\caption{CNN boost training}  
	\label{Alg:LocalReassembly}  
	\begin{algorithmic}  
		\REQUIRE{The training data $\mathcal{X,Y}$, the number of learners $K$}
		\ENSURE{The compatibility detector/classifier $f(x)$}
		\STATE Initialize training data weight $D_0=(w_{01},w_{02}, ...,w_{0n})$, where $w_{0i}=\frac{1}{n}, i=1,2,...,n$.
		\FOR{$k=1$ to $K$} 
		\item Train a network learner $G_{k}^*(x)$ to optimize Eq.~(\ref{Eqn:best_G_CNN}).
		\item Convert to discrete classification result, using Eq.~(\ref{Eqn:convert}).
		\item Calculate $\alpha_k^*$, using Eq.~(\ref{Eqn:best_a}).
		\item Update the weight distribution on training data  $D_k=\{w_{k, 1}, w_{k, 2}, ..., w_{k, n}\}$ using Eq.~(\ref{Eqn:w}).
		\ENDFOR
		\STATE The final classifier is $f(x) = \sum_{k=1}^{K} \alpha_k G_k^*(x)$
	\end{algorithmic}  
\end{algorithm}  	

\section{Global Composition Maximizing Loop Consistency}
\label{Sec:GlobalDetail}
\begin{figure}[h]
	\centering
	\begin{tabular}{ccc}
		\includegraphics[height=0.09\textheight]{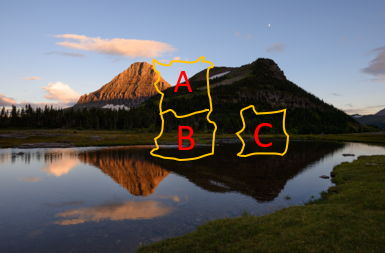} &
		\includegraphics[height=0.09\textheight]{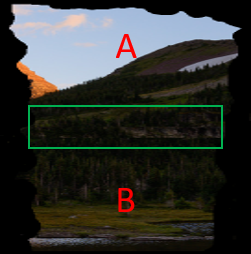} &
		\includegraphics[height=0.09\textheight]{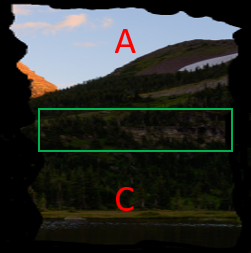}\\
		(a) Original image & (b)correct & (c) ambiguous  \\	
	\end{tabular}
	\caption{Local Ambiguity. (a) shows the original image. The correct alignment is shown in (b); but the incorrect stitching in (c) also demonstrates good pairwise compatibility according to the detector. Such local ambiguity needs to be eliminated with the help of a global composition, using mutual consistency from multiple fragments. \label{Fig:ambiguity}}
\end{figure}

After pairwise compatibility measurement, a major part of incorrect pairwise alignments have been filtered out. 
But between many fragment pairs, we still preserve more than one potential alignments. 
This is because (1) the trained compatibility classifier has not yet reached perfect accuracy, and (2) there is pairwise alignment ambiguity that can not be ruled out locally. 
Fig.~\ref{Fig:ambiguity} illustrates such an example. 
Both alignments in Fig.~\ref{Fig:ambiguity} (b) and (c) seem to produce natural stitching. 
Therefore, setting a too high threshold to strictly reject alignments (or even just keep one alignment per pair) may not be a good idea.
Instead, we keep several pairwise alignments between a fragment pair, then handle their pruning through this global composition by enforcing groupwise consensus. 

Most existing global composition algorithms adopt certain types of greedy strategies such as the best-first, spanning-tree growing, or their variants~\cite{zhang2014graph, pomeranz2011fully, gallagher2012jigsaw}, if incorrect/ambiguous alignments have higher matching or compatibility scores and are picked to occupy the positions that belong to other correct pieces, the final composition will fail because of such a local minimum. 

Loop closure has been widely adopted as a global consensus constraint in SLAM~\cite{mur2017orb}, 3D reconstruction ~\cite{choi2015indoor,le2018sparse3d}, and global structure-from-motion~\cite{cui2015global,zhu2017parallel}, and has demonstrated effective in these tasks. 	
Here, we develop two new strategies to enforce the global loop closure constraints and prune incorrect pairwise alignments.

We call the first strategy as Greedy Loop Closing (GLC). Instead of performing traditional greedy selection on edges (alignments), the greedy selection is conducted in the level of loops. Therefore, high-score edges that violates loop closure will not lead to local minima, and GLC is more robust than existing edge-based searching algorithms.
Furthermore, we also develop a second strategy, called Hierarchical Loop Merging (HLM). Instead of greedily selecting closed loops like GLC does, the decision will be made after hierarchical merging operations. Therefore, incorrect local loops that are closed but incompatible with other big loops will not lead to local minima, and HLM is more robust than GLC in solving complicated puzzles.

\subsection{Terminologies and Formulations}
\label{Sec:Terminologies}
We use a \textbf{directed multi-graph} $G=\{\mathcal{V}, \mathcal{E}\}$ to store all the image fragments and pairwise alignment candidates. 
Each vertex $v_i \in \mathcal{V}$ corresponds to an image fragment and a 2D rigid transformation matrix, or pose, $X_i \in \mathcal{X}$. 
Between each pair of vertices $(v_i, v_j)$, there are one or more edges.
Each such edge $e_{i,j,k} \in \mathcal{E}$ corresponds to a pairwise alignment, where $i,j$ are vertex indices and $k$ indicates the $k$-th potential alignments between them. 
Every edge $e_{i,j,k}$ is associated with a 2D rigid transformation matrix $T_{i,j,k}$, stitching fragment $i$ to fragment $j$. For each $T_{i,j,k}$ we have a compatibility \emph{score} $\gamma$, which is the output of the CNN classifier defined in the last section, indicating the probability of its correctness. 
Many loops $\{l_1,l_2,...,l_t\}$ can be found in graph $G$. 
A loop closure constraint is formulated on a loop $l_t$ as 

\begin{equation}
\prod_{(i,j,k)\in l_t} T_{i,j,k}= I
\end{equation}
where $I$ is the identity matrix.
A loop that satisfies this constraint is called a \emph{closed loop}.
Note that, while each edge $e_{i,j,k}$ is directed, a loop could contain it in its reversed direction $e_{j,i,k}$. In that case, we shall use its reversed transformation $T_{j,i,k}=T_{i,j,k}^{-1}$ in evaluating the loop closure constraint. 
In the following, without causing ambiguity, we may simplify the discussion of loop closure on an \textbf{undirected graph}.

\begin{figure}[h]
	\centering
	\begin{tabular}{cc}
		\includegraphics[height=0.11\textheight]{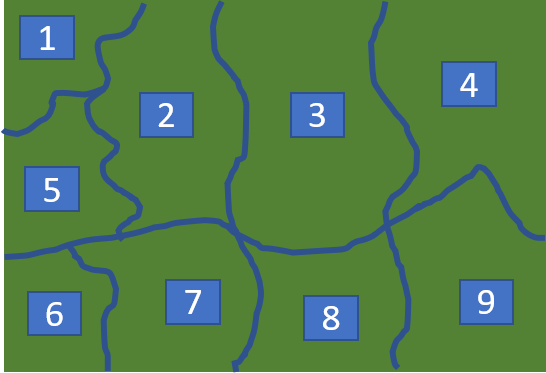} &
		\includegraphics[height=0.11\textheight]{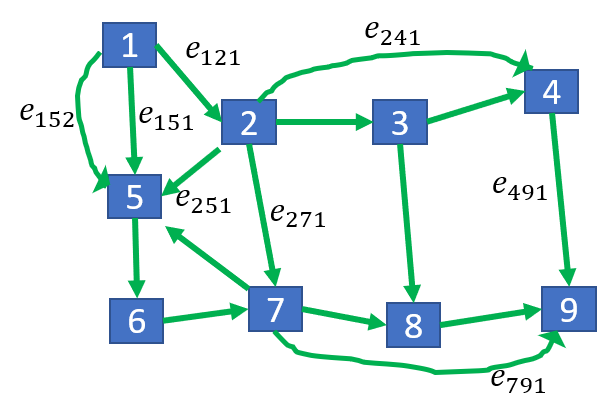} \\
		(a) & (b) \\	
	\end{tabular}
	\caption{Loop closure constraints on a directed multi-graph. 
		(a) A simulative jigsaw puzzle with 9 fragmented pieces. 
		(b) The corresponding graph model is a directed multi-graph. 
		\label{Fig:loop}}
\end{figure}
Fig.~\ref{Fig:loop} illustrates an example of simple multi-graph. 
There are multiple small loops whose lengths are $3$ or $4$, such as $(1\to 2 \to 5 \to 1)$, $(2 \to 3 \to 8 \to 7 \to 2)$, $(2 \to 7 \to 6 \to 5 \to 2)$. 
If $T_{121}*T_{251}*T_{151}^{-1}=I$, then those transformations on $(1\to 2 \to 5 \to 1)$ form a closed loop, and we consider this group of transformations to be \emph{mutually consistent}. 
The alignments that satisfy such a loop closure constraint are considered more reliable than those individual pairwise alignments receiving high local alignment scores.
This loop closure constraint provides a more global and reliable measure over local pairwise compatibility measures.

\textbf{Induced loops and mergeable loops.}
A loop is called a \emph{hole} or an \emph{induced loop}, if no two vertices of it are connected by an edge that does not itself belong to this loop.
In Fig.~\ref{Fig:loop} (b), $(2 \to 5 \to 7 \to 2)$ and $(7 \to 5 \to 6 \to 7)$ are two induced loops. 
$(2 \to 5 \to 6 \to 7 \to 2)$ is not an induced loop because $(5 \to 7)$ is connected through a path (edge) that does not belong to this loop. 
If one common edge $e_{i,j,k}$ can be found in closed loops $l_p$ and $l_q$, then $l_p$ and $l_q$ are \emph{adjacent} or \emph{mergeable}. 
$(2 \to 5 \to 7 \to 2)$ and $(7 \to 5 \to 6 \to 7)$ are mergeable because $(5 \to 7)$ is their common edge. 
(As mentioned above, we consider this on an undirected graph to simplify the notation without causing ambiguity). 
Merging induced loops results in more complicated loops.  

\textbf{Composition with Loop Closures.}
Based on the above definitions, we can formulate the global composition as an optimization problem in Eq. (\ref{Eq:GlobalObjective}).
\begin{equation}
\label{Eq:GlobalObjective}
\begin{split}
E(\mathcal{X}, &\mathcal{U}) = \min  \sum_{i,j,k} u_{i,j,k} f(X_i, X_j, T_{i,j,k}) + w_{i,j,k}(1-u_{i,j,k}) \\
& s.t. \quad \forall u_{i,j,k} \in \{0, 1\}, \; \text{and no fragment intersection}
\end{split}.
\end{equation}
where $u_{i,j,k} \in \mathcal{U}$ is an indicator variable: $u_{i,j,k} = 1$ means edge $e_{i,j,k}$ and associated transformation $T_{i,j,k}$ are selected, and $u_{i,j,k} = 0$ otherwise. $w_{i,j,k}$ is a penalty weight. 
$f(X_i, X_j, T_{i,j,k})$ measure the inconsistency between a selected pairwise alignment $T_{i,j,k}$ and the final poses $X_i, X_j$ on the nodes. 
Specifically, it can be formulated using a nonlinear least-square function \cite{grisetti2010tutorial}, 
\begin{equation}
\label{Eq:f_detail}
f(X_i, X_j, T_{i,j,k}) = e(X_i,X_j,T_{i,j,k})^T \Omega_{ij} e(X_i,X_j,T_{i,j,k})
\end{equation}
where $e(X_i,X_j,T_{i,j,k}) = \phi \left[ T_{i,j,k}^{-1}X_i^{-1}X_j\right]$ and the operator $\phi$ converts a $3\times3$ transformation matrix to a \emph{3}-dimensional vector representing the translation and rotation. If $T_{i,j,k}^{-1}X_i^{-1}X_j$ is identity, then the output is a zero vector. $\Omega_{ij}$ is a $3\times3$ weight matrix.

The objective function of Eq.~(\ref{Eq:GlobalObjective}) is defined based on the following intuition. 
Our goal is to solve pairwise alignments selection $\mathcal{U}$ and all of image fragments pose $\mathcal{X}$.
In jigsaw puzzle solving, correct pairwise alignments are always compatible with each other, while incorrect ones are prone to produce pose violations.
In other word, if an incorrect alignment is selected, it will bring more inconsistencies than a correct alignment. 
Big pose inconsistency will be reflected by a big error value of $f(X_i, X_j, T_{i,j,k})$. In this case, to minimize terms $u_{i,j,k} f(X_i, X_j, T_{i,j,k}) + w_{i,j,k}(1-u_{i,j,k})$, we tend to discard this edge and get rid of $f(X_i, X_j, T_{i,j,k})$ by setting $u_{i,j,k}=0$ and accept the penalty weight $w_{i,j,k}$. 
Therefore, selecting incorrect alignments will bring more penalties than selecting correct alignments. 
Minimizing Eq. (\ref{Eq:GlobalObjective}) is equivalent to select as many mutually consistent alignments as possible. The loop closure constraint is implicitly included in Eq.~(\ref{Eq:GlobalObjective}).
Since the edges/alignments are consistent in a closed loop, the optimization of Eq. (\ref{Eq:GlobalObjective}) can be seen as finding edges/alignments to maximize the number of \emph{compatible} loops.

\subsection{A Greedy Loop Closing (GLC) Algorithm}
\label{Sec:Greedy_Loop_Closing}

Problem (\ref{Eq:GlobalObjective}) is highly non-linear and has many local minima. 
Finding its global optimal solution is essentially NP-hard~\cite{Zhang15ICCV}. 
When the directed multi-graph $G$ is complicated, enumerating all the possible solutions is prohibitive. Hence, developing an algorithm to find an approximate solution is a more effective strategy in practice. 

In most SLAM and image reconstruction problems, registration between consecutive frames are mostly reliable, and loop closure is mainly used to refine the poses and suppress accumulative error.  
Therefore, loop closures are often formulated on a simple graph, and enforced through a best-first greedy strategy, sometimes followed by pose-graph optimization post-processing~\cite{kummerle2011g}. 
However, in fragment reassembly, a big portion of computed pairwise alignments are outliers, hence, most existing solvers are prune to local minima and often fail in composing complicated puzzles. 

Unlike existing greedy strategies, which iteratively select the best \emph{edge} that satisfies loop closure and intersection-free constraints, in this algorithm, we iteratively search for loops and fix each found one if it is closed and introduces no inter-fragment intersection.  
Specifically, the \emph{loop searching} routine is done through a Depth-First Search (DFS). 
It starts from a random edge, and randomly grows by merging adjacent edges, as long as no inter-fragment intersection is detected, until a loop $l$ is found.
Then we check whether $l$ satisfies the loop closure constraint and the intersection-free (between fragments from $l$ and fragments that are already fixed) constraint. 
If $l$ satisfied both constraints, then we \textbf{fix} this loop by selecting all the edges on this loop (by setting indicators to $1$) and discarding all their conflicting~\footnote{Two different edges between a same pair of nodes, $e_{i,j,k}$ and $e_{i,j,h}$ are conflicting, because between each pair of nodes, at most one pairwise alignment could be selected} edges (by setting indicators to $0$). 
If $l$ violates any of the two constraints, then $l$ is invalid, and will be ignored.

We keep performing this loop searching and fixing acceptable loops (selecting their loops), until (1) all the nodes in $G$ have been connected by \emph{selected edges}, or (2) the DFS search has no edge to select, or (3) a maximal searching step $N$ is reached.  
After loop searching, if there are nodes that are not connected with fixed edges through loops, we greedily select highest-score and intersection-free edge from the left undecided edges to connect them. 
Finally, we can calculate all the fragments' poses $\mathcal{X}$ using the selected edges/alignments in the final graph. 



This greedy strategy is usually efficient because any intersection-free closed loop will be fixed and related conflicted edges will be discarded once a valid loop is found. 
Compared with existing various best-edge first selection strategies adopted in existing reassembly literatures, this algorithm is less sensitive to local minima caused by single pairwise alignments that have high compatibility score but are incorrect.

However, if incorrect pairwise alignments also form a closed loop, then this strategy may get trapped locally again. 
If an incorrect-alignments-formed closed loop is selected first and its associate fragments occupy the positions where the correct closed loops should locate, then the reassembly will be incorrect in these regions. 	
Fig. \ref{Fig:global_comparison}~(a) illustrates a failed example of this greedy loop closing algorithm. 
Here the incorrect closed loop $(5 \to 8 \to 9 \to 5)$ was detected before the correct one $(4 \to 6 \to 8 \to 9 \to 4)$, and the incorrect loop occupied the positions that belong to the correct loops. The correct loop is then discarded and this leads to an incorrect reassembly. 

To further improve the robustness of the global composition, we also design another algorithm through a hierarchical loop merging strategy. 

\subsection{A Hierarchical Loop Merging (HLM) Algorithm}
\label{Sec:LoopMerging}

\begin{figure}[h]
	\centering
	\begin{tabular}{cc}
		\includegraphics[height=0.12\textheight]{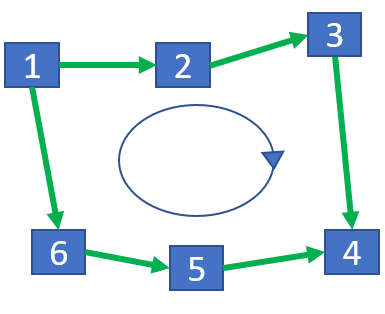} &
		\includegraphics[height=0.12\textheight]{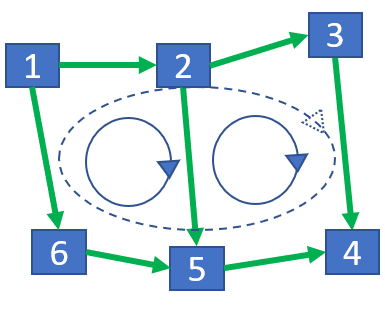} \\
		(a) Single long induced loop & (b) Merged loops \\	
	\end{tabular}
	\caption{The difference between merged loop and single long induced loop. Although both of them have same vertex, merged loop has one more compatible edge/alignment ($2\to5$). When more loops are merged, more interlocking edges/alignments need to be satisfied. Therefore, the composition reliability will be enhanced.  \label{Fig:MergedLoopComparison}}
\end{figure}

In this strategy, instead of directly fixing a found closed loop, we keep all closed loops we found, and perform the selection through an iterative merging. 
Since the true closed loops are always mutually compatible and false closed loops lead to violations, the correct solutions can be found by merging operation. 
As Fig. \ref{Fig:MergedLoopComparison} showed, the merging operation will further check the alignments/edges compatibility, and thus the composition reliability will be fortified. 
The correct probability will significantly improve with more loops merged. 


To merge as many closed loops as possible, the algorithm undergoes a \emph{bottom-up} merging phase then a \emph{top-down} merging phase.




\subsubsection{Bottom-up Merging}

We start with small \emph{induced loops}, whose lengths are $3$ or $4$. 
The set of loops found in this initial step is denoted as $\mathcal{L}^0$. 
From $\mathcal{L}^0$, we try to search loop pairs that are \emph{mergeable} (see Sec.~\ref{Sec:Terminologies} for definitions of these terminologies). 

We say two mergeable loops $l_p$ and $l_q$ are  \emph{incompatible}, if after merging, any of the following conditions are violated:
\begin{itemize}
	\item \textbf{Condition 1 (C1)} (Pose Consistency): If any vertex $v$ is in both $l_p$ and $l_q$, then the pose of $v$'s associated fragment, derived either from $l_p$ or $l_q$, should be consistent. 	
	\item \textbf{Condition 2 (C2)} (Intersection-free): For any two different vertices $v_{p} \in l_p$ and $v_{q} \in l_q$, their derived poses should not make the associated fragment overlap with each other. 
\end{itemize}  
When two mergeable loops satisfy both \textbf{C1} and \textbf{C1}, we can merge them into a bigger loop. This will result in a valid composition locally. 
If two mergeable loops violates one of these conditions and are \emph{incompatible}, then merging them leads to an invalid composition. This means at least one of these two loops are incorrect. 

\begin{figure}[h]
	\centering
	\includegraphics[height=0.18\textheight]{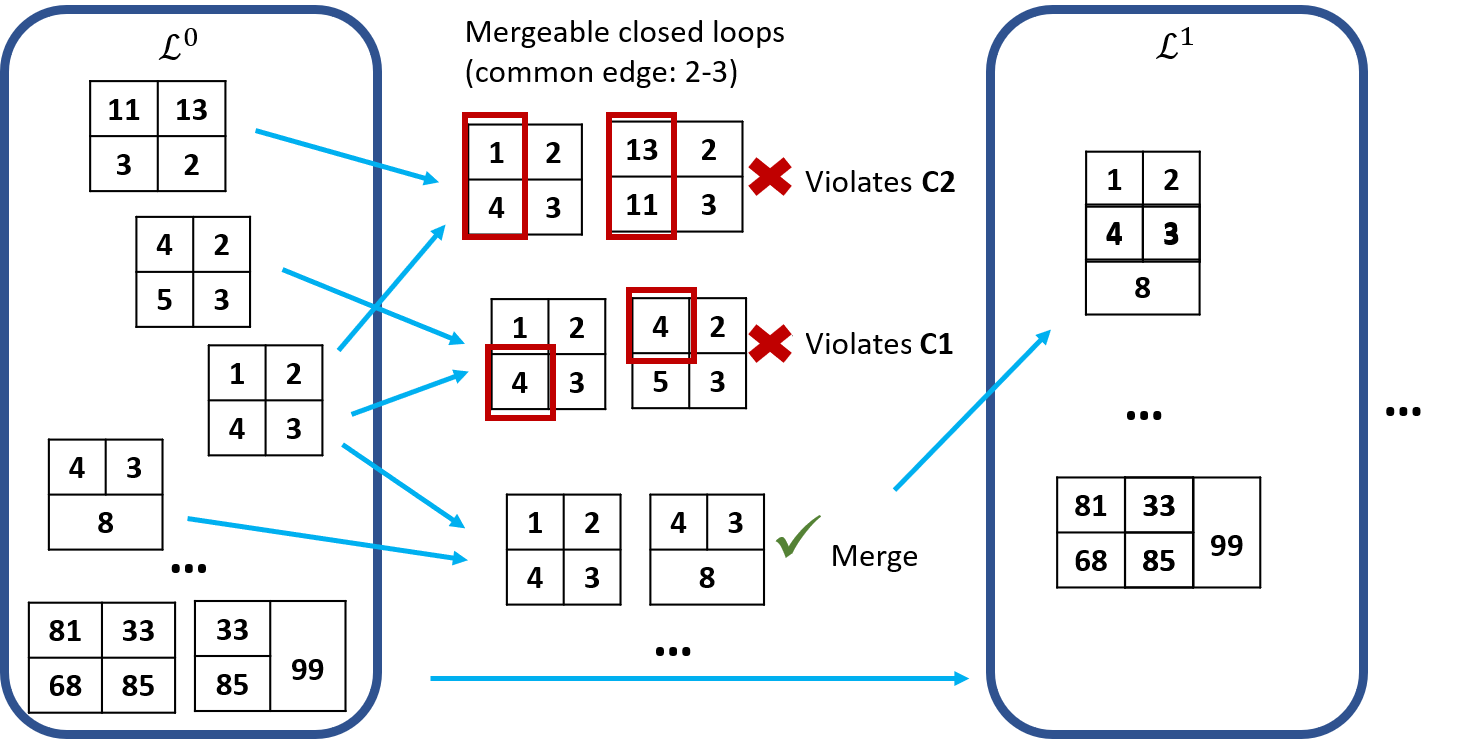}
	\caption{
		Bottom-up merging. For convenience, we use small squares to represent image fragments whose actual shapes are irregular.
		In each iteration, we try to merge all mergeable loop pairs in $\mathcal{L}^k$ and add the merged bigger loop into $\mathcal{L}^{k+1}$. 
		Merged bigger loops have higher probability to be correct than those smaller loops. \label{Fig:BottomUpMerging}}
\end{figure}

We merge loops from $\mathcal{L}^0$, and add the new merged loops into a new set $\mathcal{L}^1$. 
Then, iteratively we repeat this procedure to get $\mathcal{L}^2$, $\mathcal{L}^3$, $\ldots$, $\mathcal{L}^n$ until no more loops can be merged. 
With the growth of the \emph{compatible} loops, the probability of these big loops being correct significantly increases.  
In the last loop set, $\mathcal{L}^n$, we select a loop that has the highest sum of score and denote it as $l^*$. 
$l^*$ corresponds to the biggest reassembled patch that we have got so far through this bottom-up merging procedure. 
Fig. \ref{Fig:BottomUpMerging} gives an illustrative example of this merging procedure.

\textbf{Controlling the Complexity.}		
When merging loops in $\mathcal{L}^i$, if we enumerate all the possible merges between every pair of loops in each $\mathcal{L}^{i}$, then the algorithm's time and space complexity will both grow exponentially. 
In a worst case, the $\beta$ closed loops in set $\mathcal{L}^{i}$ could become $\beta^2$ loops in $\mathcal{L}^{i+1}$, and then grow to $\beta^4$ in the next level. 
Therefore, to restrict this exponential growing, in each level $\mathcal{L}^{i}$, we restrict the maximal number of merges we try to be a constant number $\theta_M$ ($\theta_M$ is set to $500$ in all our experiments). 
Then, in $\mathcal{L}^{i}$, we will at most get $\theta_M$ loops (most likely, fewer than that as some merge will be unacceptable and discarded). 
From all the $\frac{\theta_M \times (\theta_M -1)}{2}$ loop pairs, we randomly consider $\theta_M$ merges.

\subsubsection{Top-down Merging}

If all the fragments (nodes) are merged into a big loop, then $l^*$ gives us the final composition.
But $l^*$ may not contain all the correct pairwise alignments: some correct alignments may not be detected (through pairwise matching) or have relatively weak compatibility, these fragments may need to be stitched onto the main components through individual edge connections. 
Therefore, we further perform a top-down merging to stitch these left-out fragments (isolated vertices) or sub-patches (sub-loops). 

The top-down merging starts with $l^*$ and first check each loop in $\mathcal{L}^{n-1}$. 
If a loop is found to be compatible with $l^*$, we will merge it to $l^*$.
Specifically, we define loops $l_1$ and $l_2$ to be \emph{valuable} to each other, if they are \emph{compatible} to each other and $l_1$ contains some vertices that are not in $l_2$. 

\begin{figure}[h]
	\centering
	\includegraphics[height=0.22\textheight]{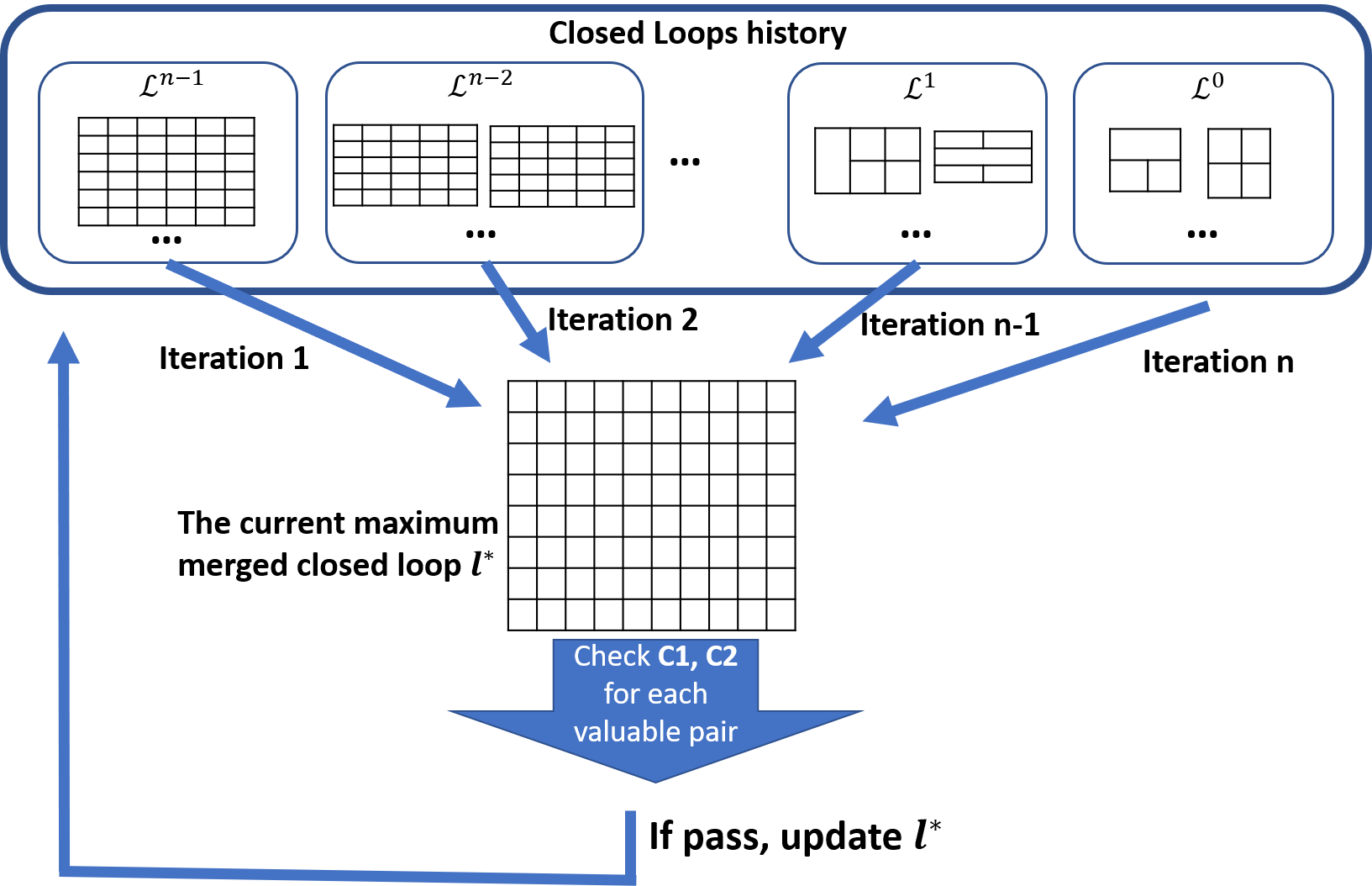}
	\caption{Top-down merging. Iteratively, we try to merge the current maximal loop $l^*$ with its valuable loops from $\mathcal{L}^{n-1}, \mathcal{L}^{n-2}, ..., \mathcal{L}^0$. 
		\label{Fig:loop_growing_top2down}}
\end{figure}

We grow $l^*$ by iteratively merging it with new valuable loops from $\mathcal{L}^{n-1}$, then $\mathcal{L}^{n-2}$, to finally, loops from $\mathcal{L}^0$.  
Fig. \ref{Fig:loop_growing_top2down} illustrates a procedure of this top-down merging. 

\textbf{Adding Left Edges after Top-down Merging.} Finally, if there are still isolated vertices, which share edges with vertices in the merged $l^*$ but are not merged. 
We then just perform a greedy growing algorithm, to iteratively pick a highest-scored edge that does not introduce fragment intersection, until no more edge can be further added.

The entire Hierarchical Loop Merging algorithm is summarized in Algorithm~\ref{Alg:HLM}.
\begin{algorithm}  
	\caption{Global Composition using HLM.}  
	\label{Alg:HLM}  
	\begin{algorithmic}  
		\REQUIRE{Multi-graph $G=\{\mathcal{V}, \mathcal{E} \}$}
		\ENSURE{An extracted simple graph $G^*$}
		\STATE $\mathcal{L}^0 \gets$ finding induced loops.
		\STATE $i=0$.
		\STATE \textbf{//Bottom-up Loop Merging}.
		\WHILE{$\mathcal{L}^i$ contains no less than $2$ loops}
		\item The number of merging $N \gets 0$.
		\WHILE{$\exists$ a mergeable pair $(l_p, l_q) \in \mathcal{L}^i$ and $N<\theta_M$}
		\IF{$(l_p, l_q)$ satisfying $\textbf{C1}, \textbf{C2}$,}
		\item Merge $l_p$ and $l_q$ into $\mathcal{L}^{i+1}$.
		\ENDIF
		\item $N \gets N+1$.
		\ENDWHILE
		\item $ i \gets i+1$.
		\ENDWHILE
		\STATE Choose the highest-score loop $l^*$ in the last set $\mathcal{L}^n$; 
		
		\STATE \textbf{//Top-down Loop Merging}.
		\FOR{$i \gets n-1 $ to $0$}
		\WHILE{$\exists$ valuable pair $(l^*, l \in \mathcal{L}^i)$ satisfying $\textbf{C1}, \textbf{C2}$, } 
		\item Merge $l$ into $l^*$.
		\ENDWHILE
		\ENDFOR
		\STATE \textbf{//Greedy Left Edge Picking for the Rest of Nodes}.				
		\STATE Sort all of edges $\{e_{i,j,k}\} \in \mathcal{E}$ from high score to low. 
		\FOR{$e_{i,j,k}$ in $\mathcal{E}$}
		\IF{$e_{i,j,k}$ connect separate vertex $v^{'}$ and $l^*$.}
		\item Merge $v^{'}$ into $l^*$.
		\ENDIF
		\ENDFOR
		\STATE Add $l^*$ to $G^*$.
	\end{algorithmic}  
\end{algorithm}

\begin{figure}[h]
	\centering                   
	\begin{tabular}{c}
		\includegraphics[height=0.22\textheight]{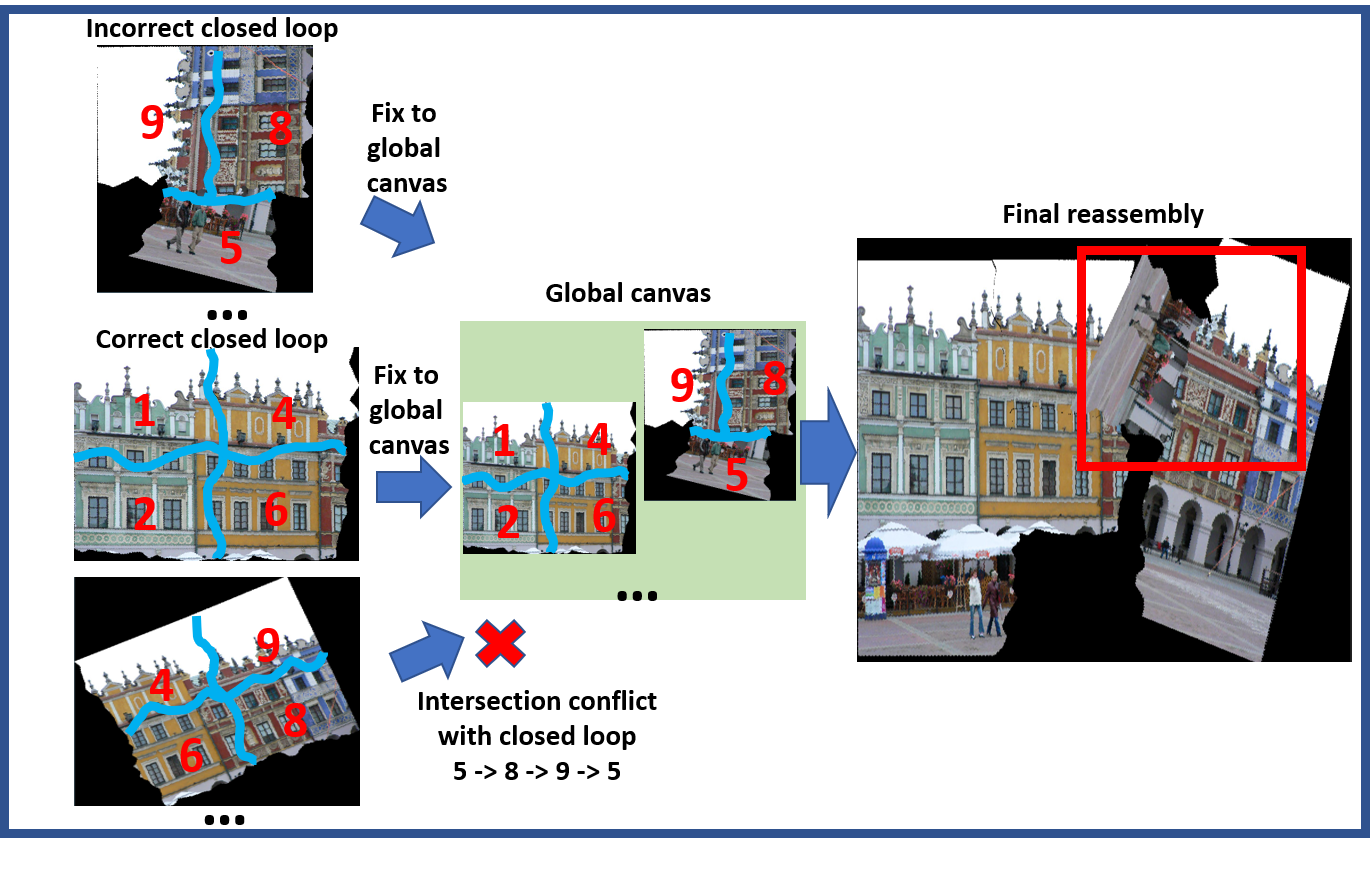} \\	
		(a) Composition by Greedy Loop Closing (GLC) \\
		\includegraphics[height=0.22\textheight]{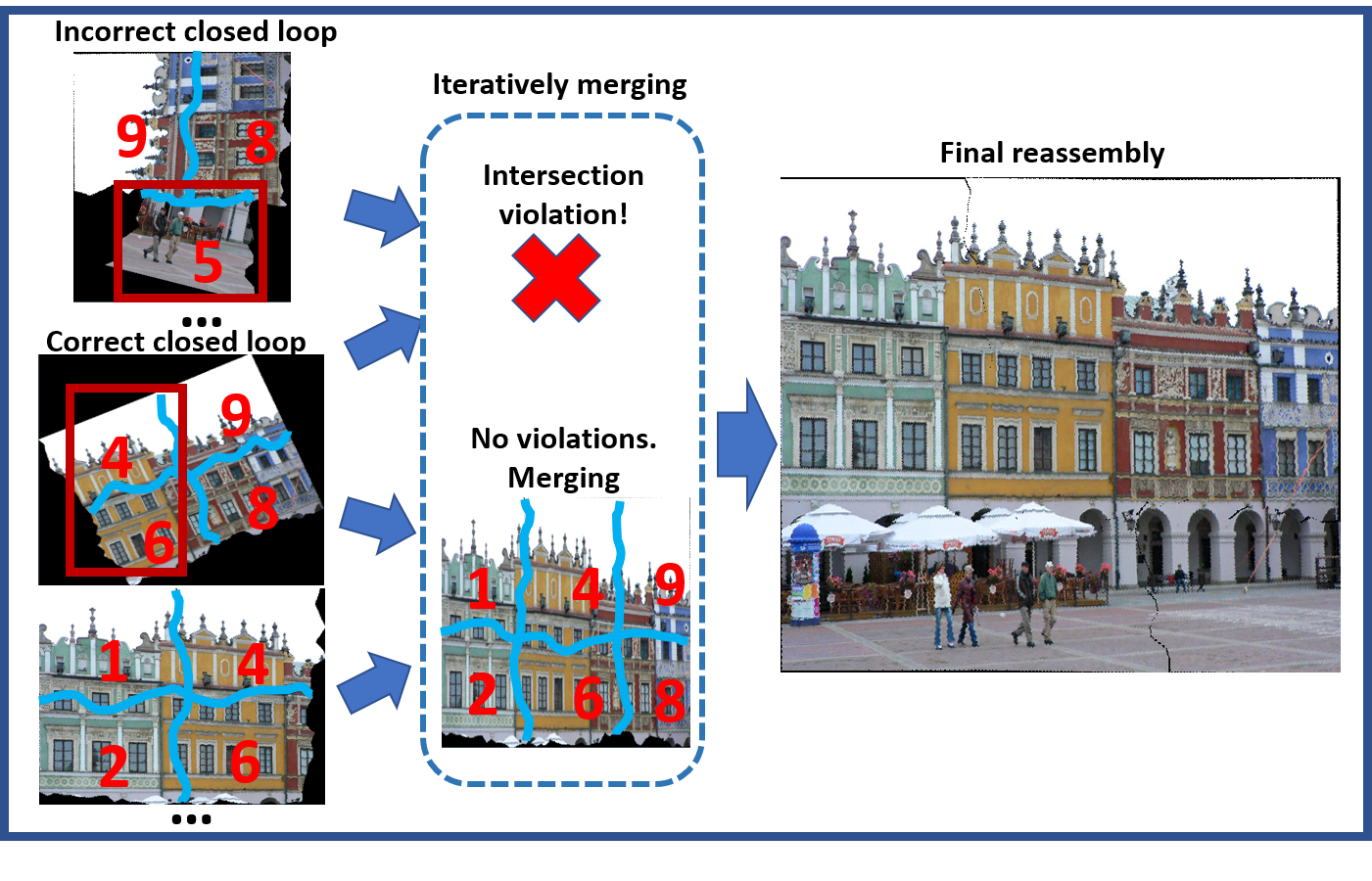}\\
		(b) Composition by Hierarchical Loop Merging (HLM) \\
	\end{tabular}
	\caption{
		GLC and HLM in Global Composition. 
		(a) With the GLC algorithm: once a closed induced loop is found, it is fixed. If such a loop is incorrect, the composition will be wrong.
		(b) With the HLM algorithm: with the same initial closed loops as (a), correct loops get merged and incorrect ones are eventually discarded as they cannot be merged. \label{Fig:global_comparison}}
\end{figure}

Fig.~\ref{Fig:global_comparison} shows an example in which the GLC algorithm (Section~\ref{Sec:Greedy_Loop_Closing}) fails but the HLM algorithm succeeds. 
With HLM, the incorrect closed loop $(5 \to 8 \to 9 \to 5)$ will be discarded, since it has conflict with the correct loop $(4 \to 6 \to 8 \to 9 \to 4)$ during merging. 
In contrast, in GLC, greedily selecting this incorrect loop leads to an undesirable local minimum and a failure in the reassembly. 

\subsubsection{Accuracy and Complexity Analysis on HLM}

The HLM algorithm aims to extract as many compatible loops as possible from the given multi-graph $G$. 
This is consistent with minimizing Eq.~(\ref{Eq:GlobalObjective}). 
Maximizing the number of compatible loops selected will minimize the second (indicator variable penalty) term, and since these loops are compatible to each other, it does not increase the first term. 
Such a merging based procedure offers a mechanism to prune \emph{false loops}, and thus, can better avoids local minima in global composition. 

\textbf{Complexity analysis.}
In Algorithm~\ref{Alg:HLM}, we use heap arrays to store vertex and edge indices, and use index sets to represent closed loops.
To find a mergeable or valuable loop pair $(l_p, l_q)$, whose lengths are $k_1, k_2$ respectively, we need $O(k_1*\lg{k_2})$ to search common elements within two heaps. 
For the same reason, the evaluation of condition \textbf{C1} can be finished in $O(k_1*\lg{k_2})$.  
The complexity of checking condition \textbf{C2} is related with image resolution, because we check the fragment intersection on a canvas. 
If the image fragment composed from a loop has $t$ pixels, then the time complexity is $O(t)$. 
Next, the loop merging operation will insert one loop's heap data structure into another, and this can be finished in $O(k_1*\lg{k_2})$. 
Since the image resolution $t >> k_1, k_2$, the complexity of a loop merging can be estimated as $O(t)$, where $t$ is the pixel number of the puzzle image. 
In our implementation, we speed up this intersection detection by implementing it using CUDA. 

In bottom-up merging, with the complexity control, the double while-loops will be run $n\theta_M$ times, where $n$ is total iteration number. 
Therefore, the total time complexity of this stage is $O(n\theta_Mt)$ (usually, n $\approx 20$ in hundreds of pieces of puzzle).
In top-down merging, we will try merging every loop in each level $\mathcal{L}^i$ with $l^*$ once. Therefore, with totally $O(n\theta_M)$ loops, the complexity is also $O(n\theta_Mt)$. 
In the final greedy selection, the time complexity is linear to the remaining edges. So if the multi-graph $G$ has $e_g$ edges, the complexity is bounded by $O(e_g t)$.
In summary, the overall complexity of HLM is $O(n\theta_M t +e_g t)$.

\section{Experiments}
\label{Sec:experiments}
We conducted experiments on two public datasets: MIT datasets \cite{cho2010probabilistic} and BGU datasets \cite{pomeranz2011fully}. However, these two datasets only contain a limited set of images. 
Popular general-purpose image databases, such as \emph{ImageNet}, are not suitable for testing jigsaw puzzle solving, because most images in these database have relatively low resolution and each often only contains a single/simple object. 
Therefore, we also create a new benchmark dataset. 
We use a website spider to automatically download images from the copyright-free website \emph{Pexels}~\cite{pexels}.
$125$ downloaded images, under different categories (e.g. street, mountain, botanical, etc), were randomly selected as training (100) and testing (25) data. 
These images are randomly cut to generate puzzles of $36$ pieces, $100$ pieces.
We denote this set of data as \textbf{TestingSet1}.
We also use $5$ additional high-resolution images to create challenging puzzles, each of which have around $400$ pieces. 
We denote this set of data as \textbf{TestingSet2}.
We have released our training and testing datasets in \url{https://github.com/Lecanyu/JigsawNet}.

\subsection{Evaluating the CNN Performance}
\begin{figure}[h]
	\centering
	\includegraphics[width=0.48\textwidth]{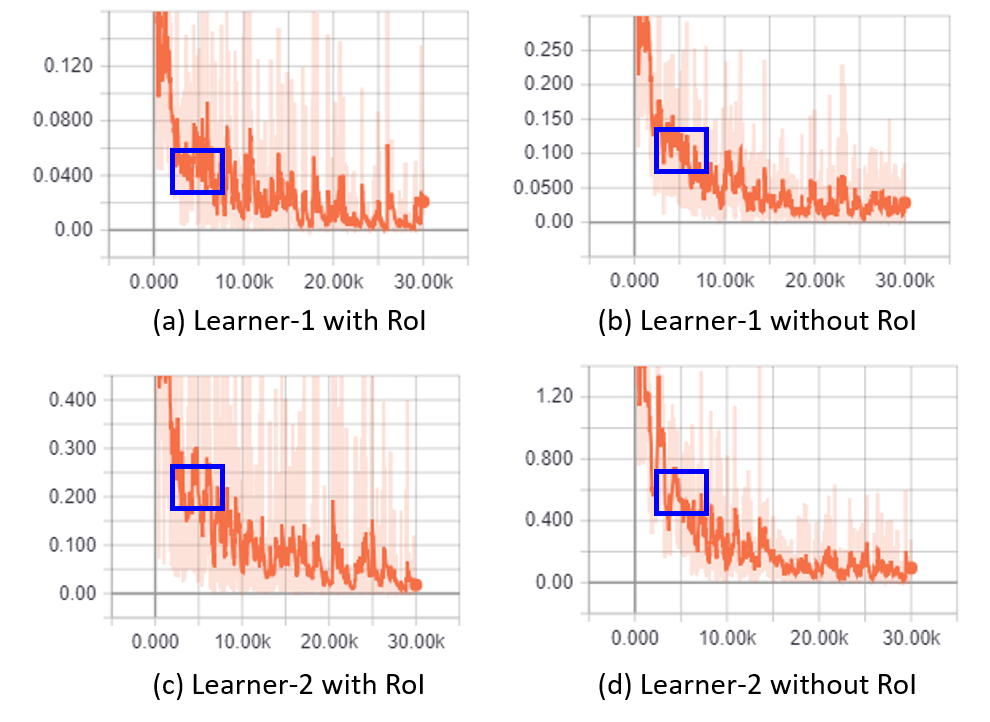}
	\caption{Convergence Comparisons of Optimization on Learners with versus without RoIs. 
		Learners with RoIs converges faster: at $5000$-th iteration (blue boxes), the loss errors of learners with RoIs are significantly smaller. Learners with RoIs also converge to smaller loss errors.
		\label{Fig:convergence}}
\end{figure}

\begin{figure}[h]
	\centering
	\includegraphics[height=0.18\textheight]{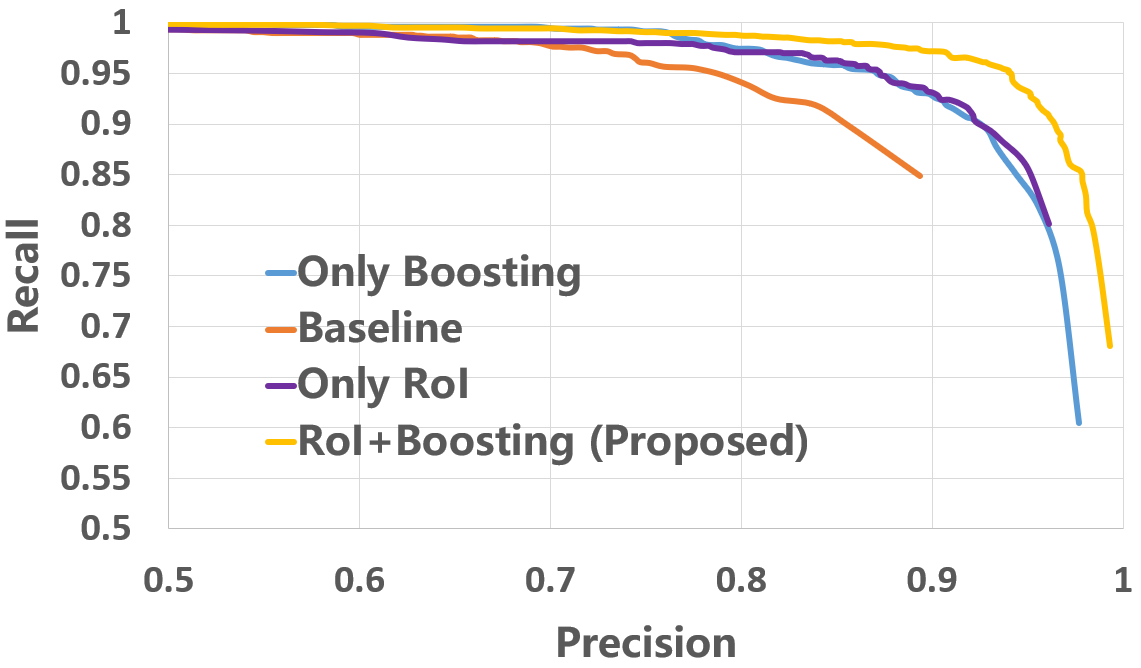}
	\caption{Precision and recall curves under different net configurations. \label{Fig:precision_recall_curve_within_net}}
\end{figure}

\begin{table}[h]
	\centering
	\caption{Classification results on different network configurations.
		TP, TN, FN, and FP represents the number of true positive, true negative, false negative, and false positive, respectively. 
		The baseline algorithm is the original network without using boosting or RoI. 
		Boosting indicates that the final classification comes from the ensemble strategy. 
		RoI means that the \emph{RoIAlign} layer is applied to replace the standard Pooling layer. 				
		All the networks in this comparisons use the same hyperparameters. \label{Tab:RoIComparison}}
	\begin{tabular}{|c||c|c|c|c|c|c|}
		\hline
		\textbf{BGU}  & TP & TN & FP & FN & Prec. & Recall \\
		\hline 
		Baseline  & 1084 & 48149 & 637 & 24 & 63.2\% & 97.8\% \\
		\hline
		Boost  & 1093 & 48404 & 382 & 15 & 74.1\% & \textbf{98.6}\% \\
		\hline
		RoI  & 1075 & 48565 & 221 & 33 & 82.9\% & 97.0\% \\
		\hline
		Boost+RoI  & 1087 & 48589 & 197 & 21 & \textbf{84.7}\% & 98.1\%  \\
		\hline
		\textbf{TestingSet1} \\
		\hline 
		Baseline  & 3848 & 202433 & 2386 & 102 & 61.7\% & 97.4\% \\
		\hline
		Boost  & 3879 & 203337 & 1482 & 71 & 72.4\% & \textbf{98.2}\% \\
		\hline
		RoI  & 3800 & 204031 & 788 & 150 & 82.8\% & 96.2\% \\
		\hline
		Boost+RoI  & 3855 & 204156 & 663 & 95 & \textbf{85.3}\% & 97.6\%  \\
		\hline
		\textbf{TestingSet2}  \\
		\hline 
		Baseline  & 2421 & 711532 & 15196 & 32 & 13.7\% & \textbf{98.7}\% \\
		\hline
		Boost  & 2396 & 721564 & 5155 & 57 & 31.7\% & 97.7\% \\
		\hline
		RoI  & 2349 & 725185 & 1534 & 104 & 60.5\% & 95.8\% \\
		\hline
		Boost+RoI & 2362 & 726049 & 670 & 91 & \textbf{77.9\%} & 96.3\% \\
		\hline
	\end{tabular}
\end{table}

\textbf{Training.}
We implemented the CNN using \emph{TensorFlow} \cite{abadi2016tensorflow} and trained it on the aforementioned 100 training images in our own dataset.  
After randomly partitioning all these images into pieces, we calculated pairwise alignments between every pair of fragments, and got around $600k$ alignments. 
These alignments, together with (1) the RoI bounding box information (which can be calculated from the alignments), and (2) a label indicating whether the alignment is correct or not (such information is available as we have the groundtruth on each image). 
We use Adam~\cite{kingma2014adam} as the optimization solver and set the batch size to $64$ and learning rate to $1e^{-4}$. The loss function is built by combining a cross-entropy term and an $l_2$-regularization term whose weight decay is $1e^{-4}$. The number of training iteration is $30k$ for every learner. The final evaluation is built up using $5$ learners. 

Using \emph{RoI} not only speeds up the training convergence but also enhances the detector's accuracy. 
Fig.~\ref{Fig:convergence} shows the convergence rates of the two learners (with versus without \emph{RoI} component). 
With \emph{RoI}, the training converges much faster and reaches smaller loss. For example, at the $5k$-th iteration (purple boxes), the losses in (a)(c) are $0.04$ and $0.23$, and the losses are $0.1$ and $0.5$ in (b)(d), respectively. 

\textbf{Testing.}
The testing is performed on the public datasets and our testing benchmarks. 
We compared the detector's performance on using different network configurations, i.e., with or without using \emph{RoI} and boosting. We calculated the \emph{recall} and \emph{precision} to evaluate the classification (compatibility detection) results.
Fig.~\ref{Fig:precision_recall_curve_within_net} shows the recall and precision curves with the four different configurations. 
Table~\ref{Tab:RoIComparison} illustrates the classification result statistics on the three testing datasets. 
Our proposed strategy that integrates both \emph{RoI} and adaptive boosting overall performs the best. It results in the best precision than the other three, and the second best recall (only slightly worse than the boosting-only strategy).

\subsection{Comparison of Pairwise Compatibility Detection with Existing Strategies}
\begin{figure}[h]
	\centering
	\begin{tabular}{cc}
		\includegraphics[height=0.11\textheight]{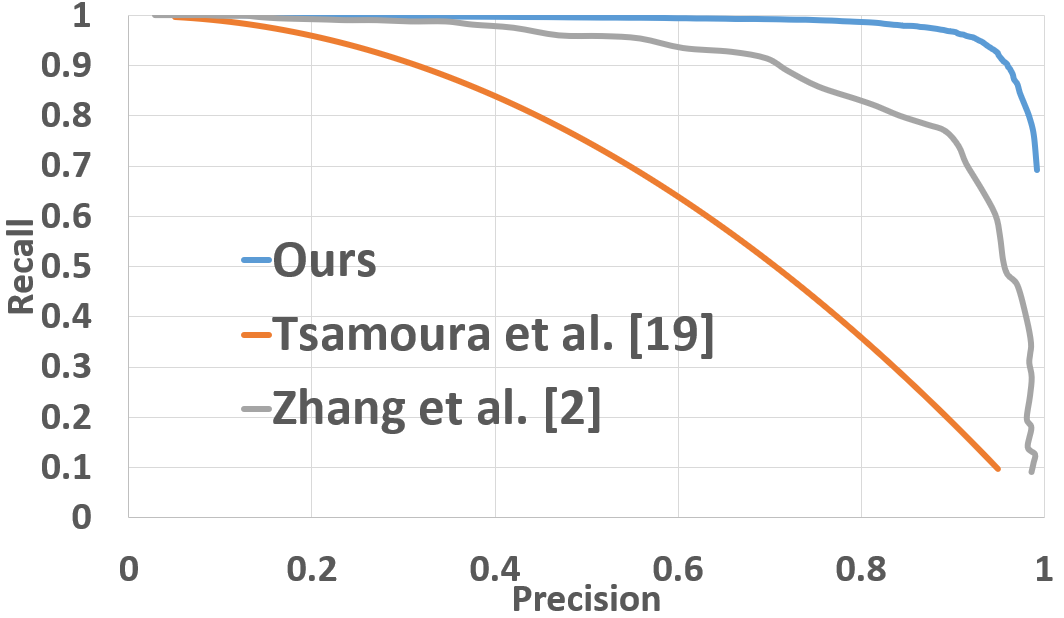} &
		\includegraphics[height=0.11\textheight]{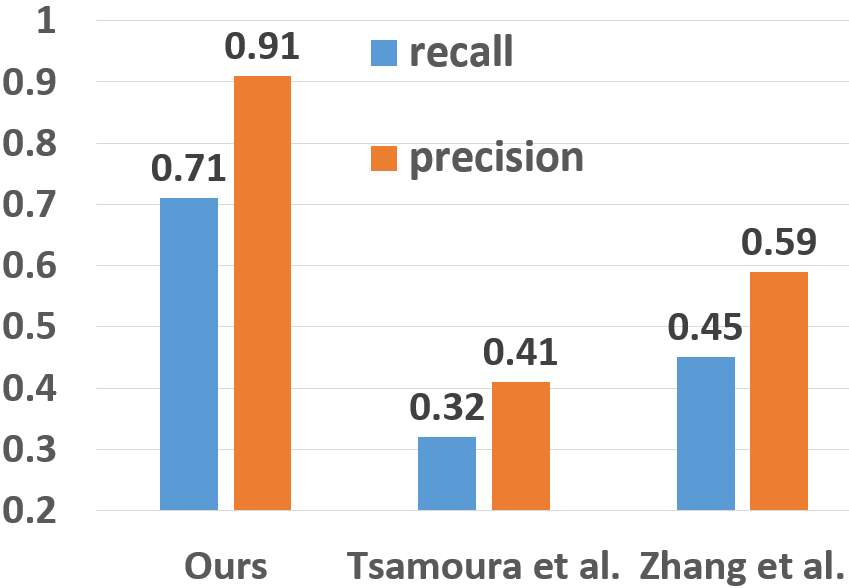}\\
		(a) & (b)  \\	
	\end{tabular}
	\caption{Pairwise compatibility measure performance on our testing benchmarks. 
		\label{Fig:precision_recall_curve}}
\end{figure}

We compared our approach with two representative methods, Tsamoura et al. \cite{tsamoura2010automatic}, and Zhang et al. \cite{zhang2014graph}.
Pairwise alignments in \cite{tsamoura2010automatic} are computed using boundary pixel color information and a longest common subsequence (LCS) algorithm.
Pairwise alignments in \cite{zhang2014graph} are computed using contour geometry and an ICP registration. 
Other matching algorithms in literatures can be considered as variants of these two approaches. 
In these algorithms, a matching/alignment score is usually produced as the output of the partial matching, and it is used to prune locally good/bad alignments. 		
We normalize these algorithms' scores to $\left[0, 1\right]$ so that they can be compared with our network's classification output. 
Then use different thresholds to draw precision and recall curves. The results are illustrated in Fig.~\ref{Fig:precision_recall_curve} (a).

We also calculate the precision and recall values in Fig.~\ref{Fig:precision_recall_curve} (b) by greedily selecting the highest score alignment to reassemble until all of fragments are connected. 
This greedy strategy can directly reflect whether the scoring mechanism is desirable.
From these experiments, we can tell that the proposed compatibility detector significantly improves the accuracy and outperforms existing scoring mechanisms in evaluating pairwise alignments. 

\subsection{Comparison of Global Reassembly Results}
Finally, but most importantly, we evaluated and compared the overall composition performance on all our testing datasets, where the puzzles have various complexity, from simple ones with $9$ pieces to complex ones that have around $400$ pieces.

In addition to the comparison with the aforementioned solvers \cite{zhang2014graph} and \cite{tsamoura2010automatic}, 
we also compared the puzzle solving results obtained from using different global composition strategies.
Specifically, first, we implement the commonly adopted best-first (BF) strategy (which iteratively picks and stitches the best non-intersecting compatible pairwise alignment). 
This strategy and its variants are commonly adopted in many existing jigsaw puzzle solvers~\cite{pomeranz2011fully, liu2011automated}. 
We compare it with our greedy loop closing (GLC) and hierarchical loop merging (HLM) algorithms. 
All these three strategies, \emph{BF}, \emph{GLC}, and \emph{HLM}, use the same CNN compatibility detector to prune raw pairwise alignments, and only differ in the global composition strategies. 
Hence, this comparison can also be viewed as an evaluation of our proposed global composition algorithms. 

To quantitatively measure the reassembly result, we consider the following metrics, some of which were also used in evaluating square puzzle solvers~\cite{cho2010probabilistic, gallagher2012jigsaw}.  
\begin{itemize}
	\item \textbf{Pose Correctness Ratio (PCR)}: the ratio of fragments whose final poses match the ground truth (i.e., the deviation is smaller than a threshold: here the rotation and shift errors are within $5^\circ$ and $100$ pixels).
	\item \textbf{Alignment Correctness Ratio (ACR)}: the ratio of correctly selected pairwise alignments in the final composition. 
	\item \textbf{Largest Component Ratio (LCR)}: the size (ratio) of the biggest recomposed component.
\end{itemize}

\begin{figure*}[!htb]
	\centering
	\begin{tabular}{cccccc}	
		\includegraphics[height=0.11\textheight]{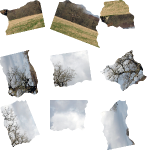} &
		\includegraphics[height=0.11\textheight]{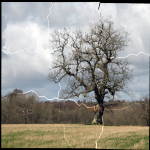} &
		\includegraphics[height=0.11\textheight]{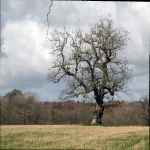} &
		\includegraphics[height=0.11\textheight]{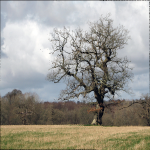} &
		\includegraphics[height=0.11\textheight]{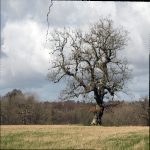} &
		\includegraphics[height=0.11\textheight]{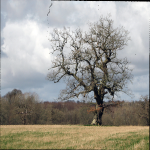} \\
		Fragments & Tsam. et al. \cite{tsamoura2010automatic} & Zhang et al. \cite{zhang2014graph} & CNN+BF & CNN+GLC & CNN+HLM  \\	
		\includegraphics[height=0.11\textheight]{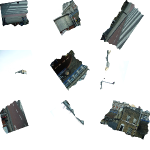} &
		\includegraphics[height=0.11\textheight]{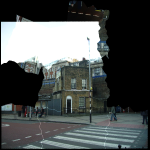} &
		\includegraphics[height=0.11\textheight]{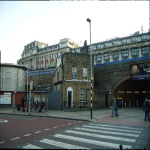} &
		\includegraphics[height=0.11\textheight]{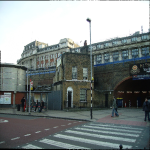} &
		\includegraphics[height=0.11\textheight]{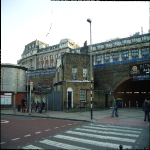} &
		\includegraphics[height=0.11\textheight]{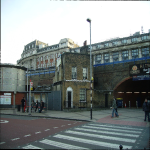} \\
		Fragments & Tsam. et al. \cite{tsamoura2010automatic} & Zhang et al. \cite{zhang2014graph} & CNN+BF & CNN+GLC & CNN+HLM  \\	
		\includegraphics[height=0.11\textheight]{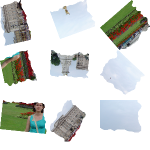} &
		\includegraphics[height=0.11\textheight]{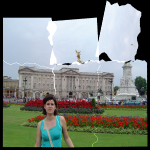} &
		\includegraphics[height=0.11\textheight]{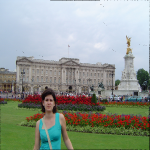} &
		\includegraphics[height=0.11\textheight]{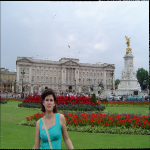} &
		\includegraphics[height=0.11\textheight]{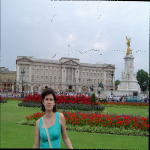} &
		\includegraphics[height=0.11\textheight]{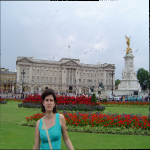} \\
		Fragments & Tsam. et al. \cite{tsamoura2010automatic} & Zhang et al. \cite{zhang2014graph} & CNN+BF & CNN+GLC  & CNN+HLM  \\	
	\end{tabular}
	\caption{
		Some reassembly results on the \textbf{MIT} data. Each puzzle contains $9$ pieces.
		\label{Fig:gallery0}}
\end{figure*}

\begin{figure*}[!htb]
	\centering
	\begin{tabular}{cccccc}
		\includegraphics[height=0.11\textheight]{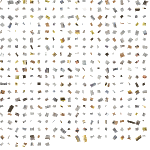} &
		\includegraphics[height=0.11\textheight]{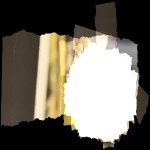} &
		\includegraphics[height=0.11\textheight]{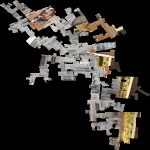} &
		\includegraphics[height=0.11\textheight]{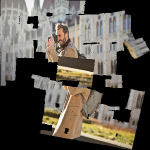} & 
		\includegraphics[height=0.11\textheight]{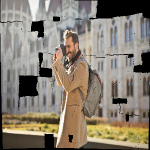} & 
		\includegraphics[height=0.11\textheight]{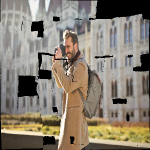} \\ 
		Fragments & Tsam. et al. \cite{tsamoura2010automatic} & Zhang et al. \cite{zhang2014graph} & CNN+BF & CNN+GLC & CNN+HLM \\	
		\includegraphics[height=0.11\textheight]{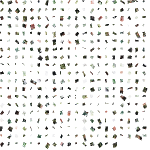} &
		\includegraphics[height=0.11\textheight]{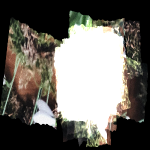} &
		\includegraphics[height=0.11\textheight]{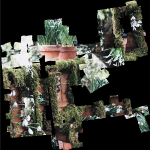} &
		\includegraphics[height=0.11\textheight]{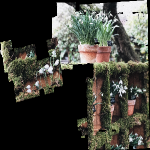} &
		\includegraphics[height=0.11\textheight]{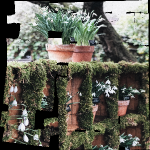} & 
		\includegraphics[height=0.11\textheight]{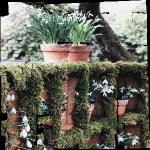} \\
		Fragments & Tsam. et al. \cite{tsamoura2010automatic} & Zhang et al. \cite{zhang2014graph} & CNN+BF & CNN+GLC & CNN+HLM \\	
		\includegraphics[height=0.11\textheight]{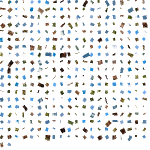} &
		\includegraphics[height=0.11\textheight]{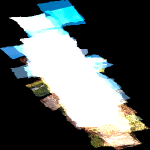} &
		\includegraphics[height=0.11\textheight]{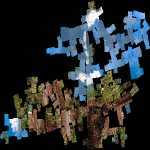} &
		\includegraphics[height=0.11\textheight]{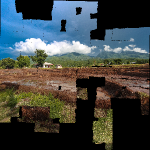} &
		\includegraphics[height=0.11\textheight]{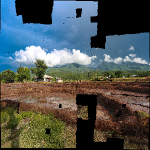} & 
		\includegraphics[height=0.11\textheight]{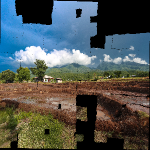} \\
		Fragments & Tsam. et al. \cite{tsamoura2010automatic} & Zhang et al. \cite{zhang2014graph} & CNN+BF & CNN+GLC & CNN+HLM \\	
	\end{tabular}
	\caption{Some reassembly results on the \textbf{TestingSet2} data. Each puzzle has around 400 pieces. \label{Fig:gallery3}}
\end{figure*}

All experiments were performed on an Intel i7-4790 CPU with GTX 1070 GPU and 16 GB RAM.	
The evaluation metrics are reported in Tables~\ref{Tab:GlobalComparisonOnMITDataset} $\sim$ \ref{Tab:GlobalComparisonOnLargeTestingDataset}.
The \emph{Avg. time} indicates the average time needed in solving each puzzle in that testing dataset.  
Some composition results are illustrated for a side-by-side comparison in Figs.~\ref{Fig:gallery0} $\sim$ \ref{Fig:gallery3}.

\begin{table}[h]
	\centering
	\caption{Reassembly on fragmented MIT datasets (totally 20 images). Each image is cut to 9 pieces. 
		CNN+BF indicates the strategy of using proposed CNN compatibility detector in alignment pruning, followed by a best first search in global composition.
		GLC and HLM stands for the greedy loop closing~\cite{le2018sparse3d} and hierarchical loop merging strategies, respectively.
		\label{Tab:GlobalComparisonOnMITDataset}}
	\begin{tabular}{|c||c|c|c|c|}
		\hline
		\textbf{MIT 9} &  PCR & ACR & LCR & Avg. time  \\
		\hline
		Tsam. \cite{tsamoura2010automatic}  &  56.7\% & - & 75.0\% &  0.82 min  \\
		\hline
		Zhang. \cite{zhang2014graph}  &  72.6\% & - & 80.3\% &  \textbf{0.76 min} \\
		\hline
		CNN + BF  &  100.0\% & 99.2\% & 100.0\% & 0.90 min  \\
		\hline 
		CNN + GLC  & 100.0\% & 99.2\% & 100.0\% &  0.91 min \\
		\hline
		CNN + HLM  & \textbf{100.0\%} & \textbf{99.2\%} & \textbf{100.0\%} & 0.94 min \\
		\hline 
	\end{tabular}
\end{table}

\begin{table}[h]
	\centering
	\caption{The overall reassembly results on BGU datasets (total 6 images). Each image is cut to 36 pieces and 100 pieces. \label{Tab:GlobalComparisonOnBGUDataset}}
	\begin{tabular}{|c||c|c|c|c|}
		\hline
		\textbf{BGU 36} &  PCR & ACR & LCR &  Avg. time  \\
		\hline
		Tsam. \cite{tsamoura2010automatic}  &  41.4\% & - & 52.8\% &  \textbf{13.73 min} \\
		\hline
		Zhang.\cite{zhang2014graph}  &  76.6\% & - & 80.3\% &  14.25 min \\
		\hline
		CNN + BF  &  99.1\% & 63.9\% & 99.1\% &  15.35 min \\
		\hline 
		CNN + GLC   & 99.1\% & 86.7\% & 99.1\% &   17.20 min \\
		\hline
		CNN + HLM  & \textbf{99.1\%} & \textbf{89.5\%} & \textbf{99.1\%} &  18.75 min \\
		\hline 
		\hline
		\textbf{BGU 100} &  PCR & ACR & LCR & Avg. time \\
		\hline
		Tsam. \cite{tsamoura2010automatic}  &  8.8\% & - & 19.6\% & \textbf{86.21 min} \\
		\hline
		Zhang. \cite{zhang2014graph}  &  33.8\% & - & 48.6\% &  88.11 min \\
		\hline
		CNN + BF &  95.0\% & 79.8\% & 93.8\% &  97.01 min \\
		\hline 
		CNN + GLC  & 96.0\% & 81.3\% & 94.2\% &  101.50 min  \\
		\hline
		CNN + HLM  & \textbf{97.2\%} & \textbf{82.8\%} & \textbf{95.3\%} & 103.03 min  \\
		\hline 
	\end{tabular}
\end{table}

\begin{table}[h]
	\centering
	\caption{The overall reassembly results on TestingSet1 datasets (total 25 images). Each image is cut to 36 pieces and 100 pieces. \label{Tab:GlobalComparisonOnTestingDataset}}
	\begin{tabular}{|c||c|c|c|c|c|}
		\hline
		\textbf{TestingSet1 36} & PCR & ACR & LCR &  Avg. time \\
		\hline
		Tsam. \cite{tsamoura2010automatic}  &  45.8\% & - & 64.9\% &   \textbf{13.48 min} \\
		\hline
		Zhang. \cite{zhang2014graph}  &  50.6\% & - & 71.3\% &  14.44 min \\
		\hline
		CNN + BF  &  96.1\% & 64.9\% & 95.7\% &  15.16 min  \\
		\hline 
		CNN + GLC  &  95.8\% & 84.0\% & 95.5\% & 17.64 min \\
		\hline
		CNN + HLM  &  \textbf{96.1\%} & \textbf{86.6\%} & \textbf{95.7\%} &  18.60 min \\
		\hline 
		\hline
		\textbf{TestingSet1 100} & PCR & ACR & LCR &  Avg. time \\
		\hline
		Tsam. \cite{tsamoura2010automatic}  &  6.5\% & - & 15.2\% &  \textbf{85.88 min} \\
		\hline
		Zhang. \cite{zhang2014graph}  &  22.3\% & - & 34.1\% &  88.96 min \\
		\hline
		CNN + BF  &  82.4\% & 72.1\% & 80.5\% &   97.92 min \\
		\hline 
		CNN + GLC &  84.1\% & 82.1\% & 80.1\% &  101.88 min \\
		\hline
		CNN + HLM  &  \textbf{86.2\%} & \textbf{88.3\%} & \textbf{81.7\%} & 103.80 min  \\
		\hline 
	\end{tabular}
\end{table}

\begin{table}[h]
	\centering
	\caption{The overall reassembly results on TestingSet2 datasets (total 5 images). Each image has around 400 pieces. \label{Tab:GlobalComparisonOnLargeTestingDataset}}
	\begin{tabular}{|c||c|c|c|c|c|}
		\hline
		\textbf{TestingSet2 400} & PCR & ACR & LCR &  Avg. time \\
		\hline
		Tsam. \cite{tsamoura2010automatic}  &  2.3\% & - & 4.8\% & \textbf{9.12 h} \\
		\hline
		Zhang. \cite{zhang2014graph}  &  11.2\% & - & 15.6\% &  10.54 h \\
		\hline
		CNN + BF  &  74.8\% & 58.6\% & 66.3\% &  11.80 h \\
		\hline 
		CNN + GLC &  \textbf{86.8\%} & 80.2\% & 85.8\% &  12.28 h  \\
		\hline
		CNN + HLM  &  86.1\% & \textbf{83.7\%} & \textbf{87.8\%} &  13.17 h \\
		\hline 
	\end{tabular}
\end{table}

We can see from these results that \cite{tsamoura2010automatic} and \cite{zhang2014graph} worked for simpler puzzles (e.g. the number of pieces is small). 
With the increase of the puzzle complexity, their performance decrease dramatically. 
In contrast, our algorithm is stable in solving these big puzzles. 
Also, with our CNN compatibility detector, even adopting simple greedy composition often produces good reassemblies, especially for small puzzles (fragments are bigger and less ambiguous). 		
Compared with the greedy strategy, the GLC and HLM strategies produce more reliable results because of the usage of loop closure constraints.

\textbf{GLC versus HLM.} 
In previous experiments, the performance difference between GLC and HLM seems small. 
This is because CNN has filtered massive incorrect alignments, the number of remaining incorrect closed loops becomes small. 
To verify this observation, we use an experiment that discards our CNN detector and only uses the pairwise matching score of \cite{Zhang15ICCV} to prune alignments in the local phase. The experiments were performed on the above MIT dataset where images were partitioned into just $9$ pieces. The results are reported in Table \ref{Tab:GlobalWithoutCNN}.
When there are many incorrect alignments, incorrect closed loops would also appear. In such scenarios, the HLM strategy outperforms the GLC algorithm. 
Therefore, when dealing with easier puzzles in which there are fewer incorrect/ambiguous alignments, GLC is suitable and it is more efficient. 
But when dealing with big and difficult puzzles, which have many small pieces and their alignments become highly unreliable, HLM offers a more robust global composition.

\begin{table}[h]
	\centering
	\caption{Reassemblies of 9-piece MIT images using GLC and HLM, but without using our CNN detector.\label{Tab:GlobalWithoutCNN}}
	\begin{tabular}{|c||c|c|c|}
		\hline
		\textbf{MIT 9} &  PCR & ACR & LCR  \\
		\hline
		Only GLC & 79.2\% & 76.1\% & 76.3\% \\
		\hline
		Only HLM  & \textbf{98.4\%} & \textbf{96.5\%} & \textbf{98.1\%}  \\
		\hline 
	\end{tabular}
\end{table}

\textbf{Comparisons with Other Puzzle Solvers.}
Without access to source/executable codes, we were not able to perform experiments using other notable solvers, such as \cite{zhu2008globally, liu2011automated}.
However, we expect these solver would perform similarly to \cite{tsamoura2010automatic} and \cite{zhang2014graph} in complex puzzles. 
Locally, these algorithms are also built upon handcrafted geometry or color based fragment descriptors and pairwise matching schemes.
In the global composition phase, they still use variants of greedy edge growing strategies, which heavily rely on the pairwise matching scores which are often ambiguous and unreliable when puzzles become complicated.

\section{Failure Cases and Limitations}
\label{Sec:Limitation}

\begin{figure}[h]
	\centering
	\includegraphics[height=0.24\textheight]{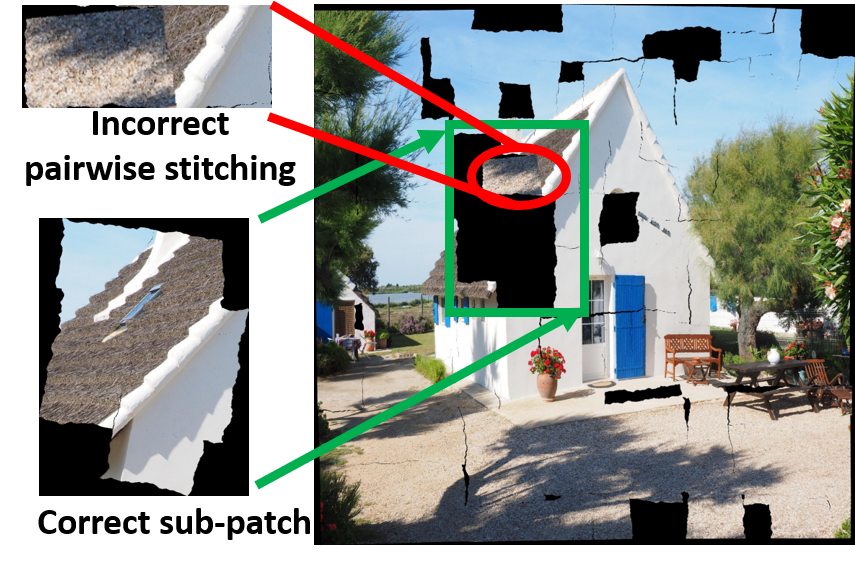}
	\caption{
		An puzzle that is incorrectly reassembled using our CNN+HLM strategy.
		Left top: the fragment that is incorrectly stitched to the main component;
		Left bottom: the correct sub-patch didn't get merged since intersection condition is violated (the incorrectly stitched fragment already took the position).
		\label{Fig:limitation}}
\end{figure}

Our global composition algorithm degenerates to greedy edge growing strategies if there are no sufficient loops. 
In such cases, when reassembling ambiguous pieces or sub-patches, the loop closure based composition could also be sensitive to the accuracy of local alignments during the greedy selection phase. 
Fig. \ref{Fig:limitation} demonstrates such a failure case. 
The correct sub-patch does not have many alignments computed with its neighboring pieces, and no loop was found to link it with the main patch. 
Without loop closure, the greedily selected alignment happens to be incorrect and a wrong piece is stitched. 

Two reasons can lead to the generation of insufficient closed loops: (1) pairwise alignment proposal algorithm misses some correct alignments. (2) The CNN misjudges some good alignments. 
To analyze them, we checked the pairwise calculation and CNN classification results. 
In this puzzle, among the total $760$ correct pairwise alignments, only $501$ ($66 \%$) were extracted.  
And within these $501$ correct alignments, $32$ were misjudged by the CNN (i.e. Recall = $93.6\%$). 
Also, among the total $153K$ extracted incorrect pairwise alignments, $152$ incorrect alignments were misjudged by the CNN (about $0.1\%$ false positive). 
Therefore, the big amount of false alignments and missed correct alignments from the pairwise alignment extraction step seems to be the major reason. 
From this analysis, we can conclude that a better pairwise alignment computation algorithm would further improve the reassembly performance.

\section{Conclusions}
\label{Sec:Conclusions}

We developed a new fragment reassembly algorithm to restore arbitrarily shredded images. 
First, we design a first CNN based compatibility detector to judge whether an alignment is correct, by evaluating whether the stitched fragment looks natural.  
Second, we developed two new global composition algorithms, GLC and HLM, to improve the composition using mutual consistency from loop closure.
With these two technical components, our algorithm has greatly outperformed the existing reassembly algorithms in handling various jigsaw puzzles.
Besides jigsaw puzzle solving, these strategies are general and could potentially be extended to other sparse reconstruction tasks.


\section*{Acknowledgments}
This work was partly supported by the National Science Foundation IIS-1320959. 
Canyu Le was supported by the National Natural Science Foundation of China 61728206, and his work was done while he was a visiting student at Louisiana State University.

\begin{figure*}[!htb]
	\centering
	\begin{tabular}{cccccc}
		\includegraphics[height=0.11\textheight]{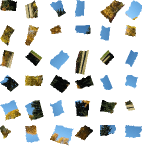} &
		\includegraphics[height=0.11\textheight]{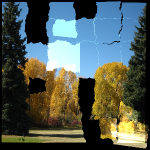} &
		\includegraphics[height=0.11\textheight]{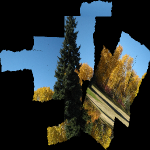} &
		\includegraphics[height=0.11\textheight]{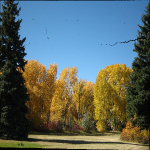} &
		\includegraphics[height=0.11\textheight]{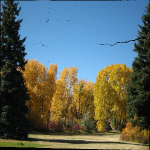} &
		\includegraphics[height=0.11\textheight]{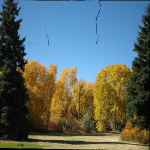} \\
		Fragments & Tsam. et al. \cite{tsamoura2010automatic} & Zhang et al. \cite{zhang2014graph} & CNN+BF & CNN+GLC & CNN+HLM \\	
		\includegraphics[height=0.11\textheight]{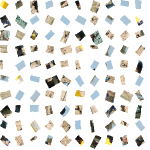} &
		\includegraphics[height=0.11\textheight]{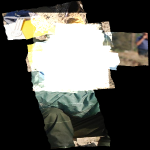} &
		\includegraphics[height=0.11\textheight]{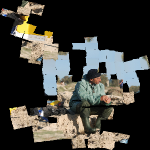} &
		\includegraphics[height=0.11\textheight]{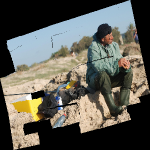} & 
		\includegraphics[height=0.11\textheight]{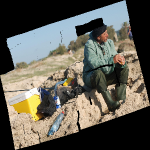} & 
		\includegraphics[height=0.11\textheight]{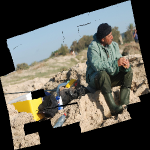} \\ 
		Fragments & Tsam. et al. \cite{tsamoura2010automatic} & Zhang et al. \cite{zhang2014graph} & CNN+BF & CNN+GLC & CNN+HLM \\
		\includegraphics[height=0.11\textheight]{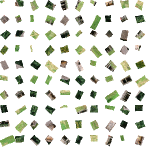} &
		\includegraphics[height=0.11\textheight]{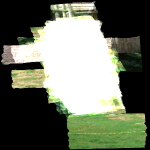} &
		\includegraphics[height=0.11\textheight]{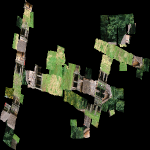} &
		\includegraphics[height=0.11\textheight]{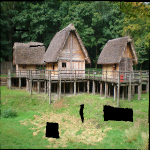} & 
		\includegraphics[height=0.11\textheight]{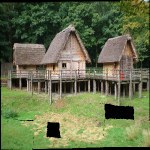} & 
		\includegraphics[height=0.11\textheight]{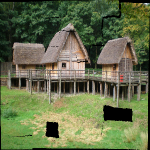} \\ 
		Fragments & Tsam. et al. \cite{tsamoura2010automatic} & Zhang et al. \cite{zhang2014graph} & CNN+BF & CNN+GLC & CNN+HLM \\
	\end{tabular}
	\caption{
		Some reassembly results on the \textbf{BGU} data. The first row shows a reassembly of a 36-piece puzzle. The second and third rows show reassemblies of a 100-piece puzzles.
		\label{Fig:gallery1}}
\end{figure*}

\begin{figure*}[!htb]
	\centering
	\begin{tabular}{cccccc}
		\includegraphics[height=0.11\textheight]{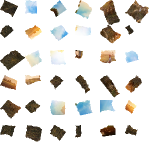} &
		\includegraphics[height=0.11\textheight]{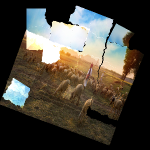} &
		\includegraphics[height=0.11\textheight]{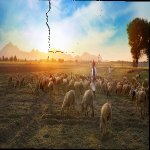} &
		\includegraphics[height=0.11\textheight]{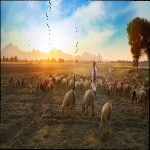} & 
		\includegraphics[height=0.11\textheight]{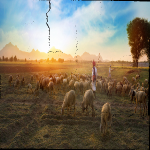} & 
		\includegraphics[height=0.11\textheight]{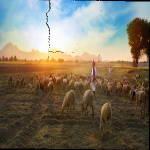} \\
		Fragments & Tsam. et al. \cite{tsamoura2010automatic} & Zhang et al. \cite{zhang2014graph} & CNN+BF & CNN+GLC & CNN+HLM \\
		\includegraphics[height=0.11\textheight]{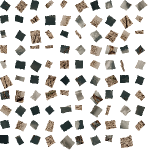} &
		\includegraphics[height=0.11\textheight]{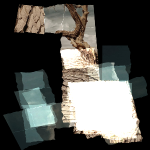} &
		\includegraphics[height=0.11\textheight]{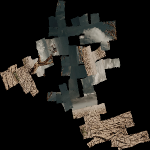} &
		\includegraphics[height=0.11\textheight]{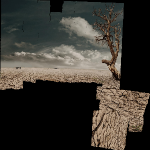} & 
		\includegraphics[height=0.11\textheight]{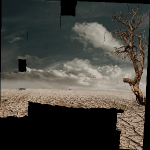} & 
		\includegraphics[height=0.11\textheight]{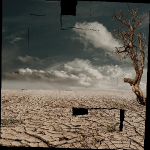} \\
		Fragments & Tsam. et al. \cite{tsamoura2010automatic} & Zhang et al. \cite{zhang2014graph} & CNN+BF & CNN+GLC & CNN+HLM  \\
		\includegraphics[height=0.11\textheight]{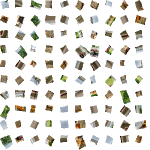} &
		\includegraphics[height=0.11\textheight]{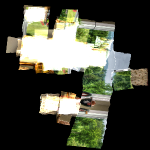} &
		\includegraphics[height=0.11\textheight]{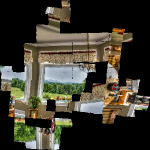} &
		\includegraphics[height=0.11\textheight]{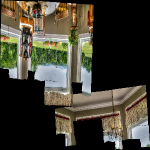} & 
		\includegraphics[height=0.11\textheight]{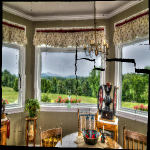} & 
		\includegraphics[height=0.11\textheight]{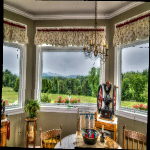} \\
		Fragments & Tsam. et al. \cite{tsamoura2010automatic} & Zhang et al. \cite{zhang2014graph} & CNN+BF & CNN+GLC & CNN+HLM \\	
	\end{tabular}
	\caption{Some reassembly results on the \textbf{TestingSet1} data. The first row shows a reassembly of a 36-piece puzzle. The second and third rows show the reassemblies of two 100-piece puzzles.
		\label{Fig:gallery2}}
\end{figure*}

\bibliographystyle{IEEEtran}
\bibliography{ref}
\ifCLASSOPTIONcaptionsoff
  \newpage
\fi

\end{document}